%% file: iclr2027_conference.tex
\documentclass{article}
\usepackage{iclr2027_conference,times}

\input{math_commands.tex}

\usepackage{booktabs}
\usepackage{multirow}
\usepackage{graphicx}
\usepackage{xcolor}
\usepackage{wrapfig}

\usepackage{hyperref}
\usepackage{url}

\newcommand{\method}[1]{\textsc{#1}}
\newcommand{\bg}{\mathrm{bg}}

\title{ActiveSAM: Fast and Accurate\\Open-Vocabulary Semantic Segmentation\\with Frozen SAM~3}

\author{%
  Tran Dinh Tien \quad Zhiqiang Shen\textsuperscript{$\dag$} \\
  VILA Lab, MBZUAI \\
  \textsuperscript{$\dag$}Correspondence: \texttt{zhiqiang.shen@mbzuai.ac.ae}
}

\iclrfinalcopy

\begin{document}

\maketitle
\lhead{Preprint}

\begin{abstract}
    Segment Anything Model~3 (SAM~3) provides a strong frozen backbone for concept-prompted segmentation, but applying it directly to open-vocabulary semantic segmentation (OVSS) is inefficient: full-resolution decoding is typically run over the entire dataset vocabulary, whereas each image contains only a small active subset of classes. We introduce ActiveSAM, a training-free inference framework that turns SAM~3 into an active-vocabulary segmenter. ActiveSAM first canonicalizes and expands class prompts, then uses evidence-proportional grounding to estimate an image-conditioned active set from a low-resolution presence preview. Only retained prompts receive full-resolution mask prediction, using bucketed prompt multiplexing with the frozen SAM~3 decoder. The preview stage uses only class-presence evidence and skips unnecessary segmentation-head computation. To resolve overlapping concept responses, exclusive concept decoding compares each pixel's joint score vector with class signatures estimated once per vocabulary from unlabeled images. ActiveSAM requires no weight updates, no oracle class-presence labels and no per-dataset hyperparameter tuning. Across eight OVSS benchmarks, ActiveSAM improves the speed-accuracy tradeoff of training-free open-vocabulary semantic segmentation, outperforming the current state-of-the-art SegEarth-OV3 by $+2.1$ mIoU on average while running much faster, with $7.3$--$12.2\times$ speedups on large-vocabulary datasets. ActiveSAM also achieves the highest accuracy under image corruptions that simulate real-world distribution shift, making it well-suited for deployment in noisy-input domains such as autonomous driving and embodied AI. Code is available at \url{https://github.com/VILA-Lab/ActiveSAM}
\end{abstract}

\section{Introduction}
\label{sec:introduction}
Real perception systems rarely operate with a fixed label set. Autonomous-driving stacks encounter new traffic objects, embodied agents receive task-specific object names, and image-analysis workflows revise their taxonomies as the task changes. In these scenarios, collecting new pixel-level annotations and retraining a segmentation model for every vocabulary update is impractical. Open-vocabulary semantic segmentation (OVSS) targets this setting: the model receives a vocabulary at test time and should produce dense masks for those classes without collecting new pixel annotations or retraining for every category update. This capability is particularly valuable as segmentation models move from closed environments to open-world deployment, where the deployment vocabulary can be large, dynamic, and only partially relevant to an individual image.

\begin{figure}[t]
    \centering
    \includegraphics[width=1.0\linewidth]{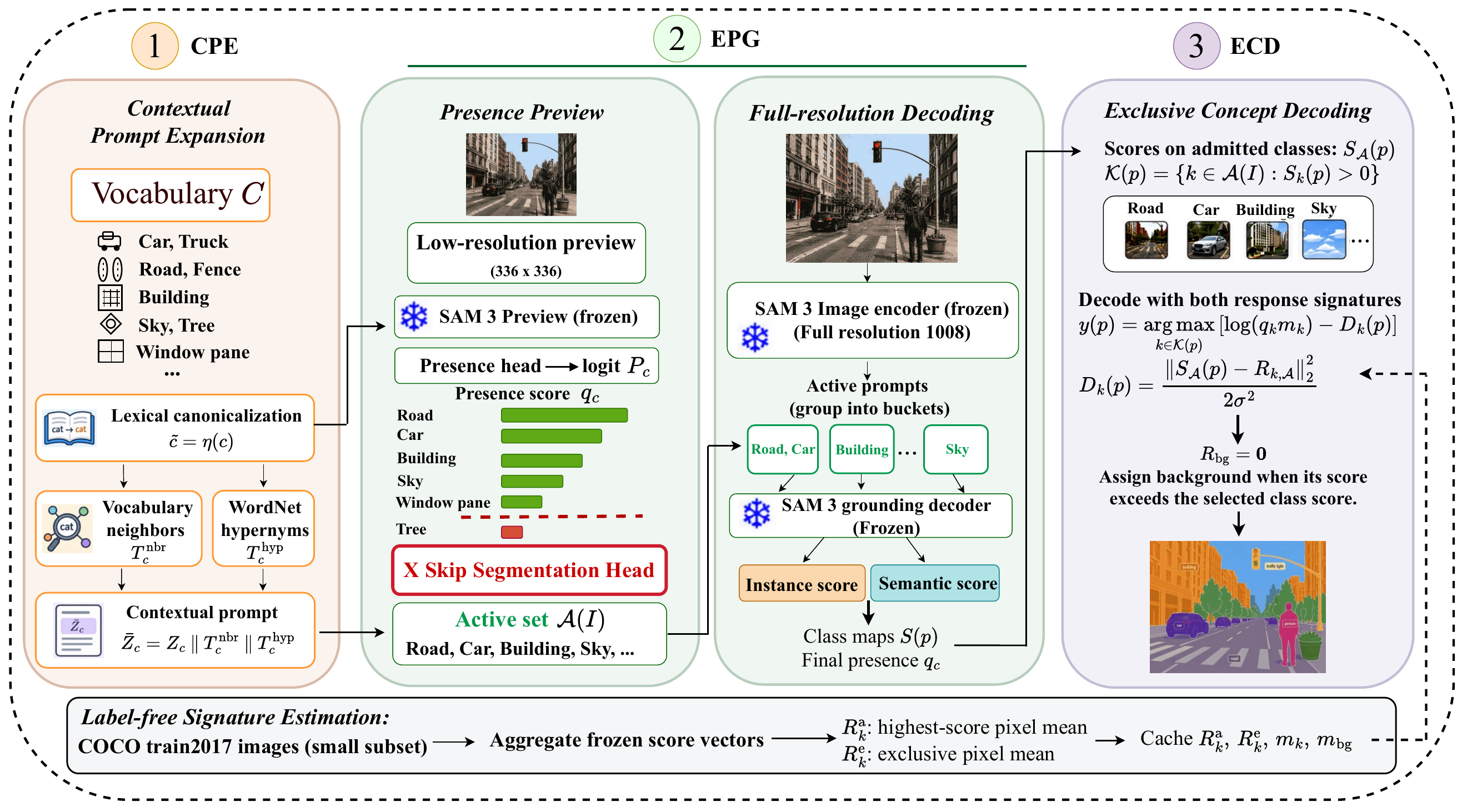}
    \vspace{-0.25in}
    \caption{Overview of ActiveSAM, which decodes only the prompts relevant to each image and resolves overlapping concept responses, with all SAM~3 weights frozen. (1) \textbf{Contextual Prompt Expansion (CPE)} canonicalizes class names and expands their prompts. (2) A low-resolution \textbf{presence preview} skips the segmentation head and admits only the classes likely to be present. (3) \textbf{Full-resolution decoding} encodes the image once and decodes the active prompts in buckets. Steps (2) and (3) form \textbf{Evidence-Proportional Grounding (EPG)}. (4) \textbf{Exclusive Concept Decoding (ECD)} labels each pixel by comparing its score vector with class signatures estimated once per vocabulary from unlabeled images.}
    \label{fig:activesam_overview}
    \vspace{-0.24in}
\end{figure}

Vision-language models have made this possible by adapting CLIP-style \citep{radford2021learning,jia2021scaling} image-text alignment to dense prediction through region classification \citep{li2022language,cho2024cat}, mask proposals \citep{ghiasi2022scaling,ding2021decoupling}, side adapters \citep{xu2023side}, diffusion features \citep{xu2023open}, and universal segmentation frameworks \citep{zhang2023simple}. Recent methods have increasingly focused on training-free OVSS, where foundation models are kept frozen and composed at inference time \citep{zhou2022extract,bousselham2024grounding,wang2024sclip,hajimiri2025pay,lan2024proxyclip,zhang2025corrclip,yang2025resclip,chen2025training}. 

\begin{wrapfigure}{r}{0.55\textwidth}
  \centering
  \vspace{-0.05in}
  \includegraphics[width=1.0\linewidth]{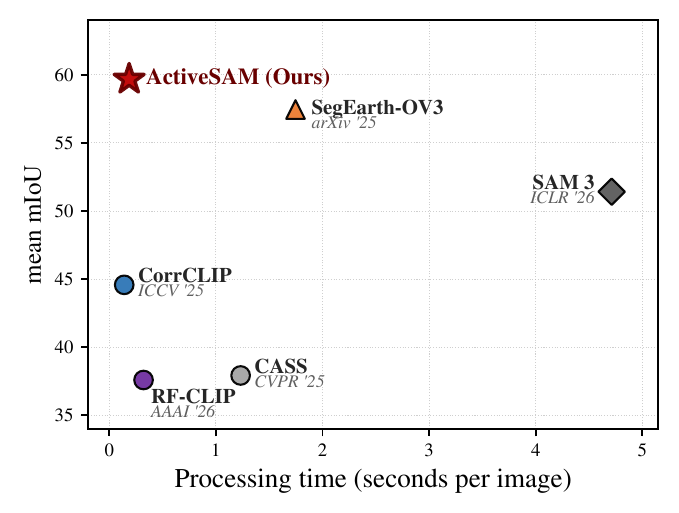}
  \vspace{-0.3in}
  \caption{ActiveSAM (\textcolor{red!70!black}{$\star$} Ours) achieves state-of-the-art mIoU while running significantly faster than SegEarth-OV3~\citep{li2025segearth}. Reported values are averaged over six benchmarks: Context59, Context60, COCO-Object, COCO-Stuff, ADE20K, and Pascal VOC21.}
  \label{fig:fps_vs_mIoU}
   \vspace{-0.1in}
\end{wrapfigure}
However, directly applying image-level recognition to pixel-level tasks often yields coarse results, as OVSS demands precise boundaries, local consistency, and fine-grained separation in complex scenes, and recent methods import spatial correspondence \citep{lan2024proxyclip}, reconstruct CLIP patch correlations \citep{zhang2025corrclip}, or modify attention \citep{wang2024sclip,hajimiri2025pay,yang2025resclip} to close this gap. This limitation has motivated a shift toward segmentation foundation models. SAM \citep{kirillov2023segment} and SAM 2 \citep{ravi2024sam} provide strong spatial masks, and recent work extends promptable segmentation with language prompts or VLM embeddings \citep{xiao2026openworldsam}. SAM~3 \citep{carion2025sam} makes the connection even more direct: it accepts concept prompts such as noun phrases, returns masks for matching instances, and decouples recognition from localization with a presence head. Recent SAM~3-based OVSS pipelines therefore convert the evaluation vocabulary into text prompts, obtain class-conditioned semantic and instance evidence, and use presence scores to suppress absent categories \citep{li2025segearth}. While this approach improves mask quality, it creates a new systems bottleneck: full-resolution prompt-conditioned decoding scales with the vocabulary, although any individual image contains only a small active subset of classes.

A second limitation concerns how concept masks are converted into semantic labels. SAM~3 evaluates each prompt separately and can give high scores to related concepts on the same pixels. A rider can match both \emph{rider} and \emph{person} because both descriptions are valid. Semantic segmentation instead requires one label per pixel. Selecting the highest score can label rider pixels as \emph{person} even when the model also responds strongly to \emph{rider}. Filtering absent classes cannot resolve this ambiguity when both classes are present. The response to \emph{rider} still provides useful evidence even when its score is lower. This motivates a decision that considers the class scores together rather than relying only on the highest score.

We propose \textbf{ActiveSAM}, a training-free inference framework that turns frozen SAM~3 \citep{carion2025sam} into an active-vocabulary segmenter. The key idea is to use SAM~3's presence signal to prune absent classes before full-resolution decoding, rather than only as a filtering score after decoding. As shown in Figure~\ref{fig:activesam_overview}, Contextual Prompt Expansion (CPE) first canonicalizes class names and adds cached semantic context. Evidence-Proportional Grounding (EPG) then selects prompts through a low-resolution presence preview that skips the segmentation head. The same presence threshold screens the selected prompts again at full resolution before mask prediction. Retained prompts are decoded in buckets that reuse image features. The number of decoded prompts therefore adapts to the content of each image, without a fixed class budget.

To distinguish overlapping concepts, we introduce Exclusive Concept Decoding (ECD). Each class is represented by \emph{response signatures} that summarize the model's scores across concepts on pixels selected for that class by the frozen model. ECD compares each pixel's score vector with these signatures using a nearest-signature classifier with a class prior. It first revises the highest-score label only when two signature estimates agree on the replacement and both the original and proposed classes have enough calibration pixels. For vocabularies with a background label, ECD then compares the selected class against a \emph{null signature}, a zero score vector representing background.

The signatures are estimated once per vocabulary from one shared pool of unlabeled COCO training images, without semantic annotations or evaluation images. ActiveSAM keeps all model weights frozen, requires no oracle class-presence labels, and uses the global set of hyperparameters across all eight benchmarks.

\textbf{Our main contributions are summarized as follows:}
\begin{itemize}
    \item We introduce ActiveSAM, a training-free inference framework that uses SAM~3's presence predictions to decide which prompts receive full-resolution mask decoding for each image, replacing the standard but inefficient practice of decoding the entire dataset vocabulary on every image. The framework combines three proposed components: \emph{Contextual Prompt Expansion} (CPE) for richer class prompts, \emph{Evidence-Proportional Grounding} (EPG) for prompt selection at two resolutions, and \emph{Exclusive Concept Decoding} (ECD) for assigning semantic labels from joint concept responses. No SAM~3 weights are updated.
    
    \item ActiveSAM achieves the highest mean mIoU among the compared methods across eight standard OVSS benchmarks. It reaches $66.02\%$ mean mIoU, outperforms the current state-of-the-art method SegEarth-OV3 \citep{li2025segearth} by $2.1$ points while dramatically running faster. On large vocabulary datasets, ActiveSAM achieves $7.3\times$ to $12.2\times$ faster while improving segmentation accuracy on each benchmark (Figure~\ref{fig:fps_vs_mIoU}). Table~\ref{tab:fps_vs_mIoU_per_dataset} provides the per-dataset comparison.

   \item ActiveSAM demonstrates strong robustness to distribution shifts. Through rigorous testing on common image corruptions (noise, blur, JPEG compression, fog), we show that ActiveSAM maintains high performance where traditional CLIP-based baselines suffer severe accuracy degradation, a crucial property for real-world applications such as autonomous driving and embodied AI.
\end{itemize}
\section{Related Work}
\label{sec:related-work}
\noindent{\textbf{Open-vocabulary semantic segmentation.}} Early OVSS methods such as \method{LSeg} \citep{li2022language} align pixel embeddings with text-label embeddings. \method{GroupViT} \citep{xu2022groupvit} shows that semantic grouping can emerge from text supervision, while \method{OpenSeg} \citep{ghiasi2022scaling} scales open-vocabulary segmentation using image/mask-level alignment. \method{ZegFormer} \citep{ding2021decoupling} decouples zero-shot segmentation into class-agnostic grouping and segment-level classification. \method{SegCLIP} \citep{luo2023segclip} learns patch aggregation with learnable centers, and \method{TCL} \citep{cha2023learning} and \method{CoDe} \citep{wu2024image} learn text-grounded masks by decomposing paired images and texts into region and word components. \method{CLIP-DINOiser} \citep{wysoczanska2024clip} uses a DINO \citep{caron2021emerging} backbone to strengthen CLIP dense features. \method{Talk2DINO} \citep{barsellotti2025talking} bridges DINOv2 \citep{oquab2023dinov2} features with CLIP language embeddings by training a lightweight alignment layer on an external dataset. Despite their strong performance, these approaches require additional training, making them less suitable for settings where additional parameter optimization is impractical.

\textbf{Training-free open-vocabulary semantic segmentation.}
Training-free OVSS aims to preserve segmentation capability without requiring learnable parameters or additional training. \method{MaskCLIP} \citep{zhou2022extract} adapts frozen CLIP for dense prediction without additional training or pixel-level annotations. Later training-free methods try to modify CLIP attention at inference time: \method{SCLIP} \citep{wang2024sclip} explores deeper self-attention layers, \method{ProxyCLIP} \citep{lan2024proxyclip} uses proxy attention, \method{NACLIP} \citep{hajimiri2025pay} enforces neighbor-aware localization, \method{CLIPtrase} \citep{shao2024explore} recalibrates patch self-correlations, \method{ResCLIP} \citep{yang2025resclip} uses residual cross-correlation self-attention with semantic feedback refinement, and \method{DIH-CLIP} \citep{duan2025dih} exploits diversity across multi-head self-attention.

Several approaches combine the CLIP-based model with other strong foundation models. \method{FreeDA} \citep{barsellotti2024training} builds offline diffusion-augmented prototypes for open-vocabulary segmentation. \method{CorrCLIP} \citep{zhang2025corrclip} reconstructs CLIP patch correlations using multiple foundation models such as DINO \citep{caron2021emerging} and SAM 2 \citep{ravi2024sam}, and \method{Trident} \citep{shi2025harnessing} combines CLIP, DINO, and SAM \citep{kirillov2023segment} for high-resolution training-free segmentation.

\textbf{SAM-based open-vocabulary semantic segmentation.}
The \method{Segment Anything Model} (\method{SAM}) \citep{kirillov2023segment} introduced promptable segmentation with strong spatial masks, and \method{SAM 2} \citep{ravi2024sam} extended this idea to images and videos, but neither solves semantic labeling over an arbitrary vocabulary. \method{SAM~3} \citep{carion2025sam} moves closer to semantic open-vocabulary segmentation by associating segmentation with concepts. \method{SegEarth-OV3} \citep{li2025segearth} explores \method{SAM~3} for remote-sensing OVSS and natural OVSS benchmarks. This method uses dual-head mask fusion and the presence head for decoder phase and generates the final labels. \method{MM-OVSeg} \citep{wei2026mm} extends open-vocabulary segmentation to multimodal remote sensing. Our ActiveSAM is related to \method{SAM~3} \citep{carion2025sam} and \method{SegEarth-OV3} \citep{li2025segearth}, but it aims at two different important questions: How can we exploit SAM~3's strong backbone while reducing the computational cost of standard SAM~3-based inference, and how can its independent concept responses be turned into one label per pixel? Instead of applying the conventional encoder-decoder pass of SAM~3, our method uses a cheap presence-only preview to select an image-specific active vocabulary before full-resolution decoding, then decodes only active class prompts with bucketed class-prompt multiplexing and return the predicted class maps jointly instead of taking their pixelwise maximum.
\section{ActiveSAM}
\label{sec:method}

\paragraph{Problem setup.}
Given an image $I$ and a semantic vocabulary $\mathcal{C}=\{c_1,\ldots,c_C\}$ containing $C$ classes, our goal is to predict a dense label map $\hat{y}\in(\mathcal{C}\cup\{\mathrm{bg}\})^{H\times W}$, where $\hat{y}$ assigns one semantic label to each pixel, $H$ and $W$ are the output height and width, and $\mathrm{bg}$ denotes background for benchmarks that define an explicit background class. We build on a frozen SAM~3 model, which performs concept-prompted segmentation and decouples recognition from localization with a presence head \citep{carion2025sam}. ActiveSAM is training-free and keeps all SAM~3 weights frozen.

\paragraph{Overview.}
\label{subsec:method-overview}
ActiveSAM consists of three components. First, Contextual Prompt Expansion (CPE) constructs and caches a contextual prompt $\bar{Z}_c$ for each class $c\in\mathcal{C}$. Second, Evidence-Proportional Grounding (EPG) uses a low-resolution presence preview to remove unlikely prompts before full-resolution decoding. The full-resolution image encoder is evaluated once, and the prompts of the active set are processed in buckets. After one decoder layer, their presence is checked again using the same global threshold $\tau$, and only prompts that still have sufficient evidence continue through the full decoder layers and the segmentation head. Third, Exclusive Concept Decoding (ECD) converts the resulting class score maps into one semantic label per pixel. Rather than selecting only the class with the highest independent score, ECD compares the full vector of class scores with cached class signatures estimated once per vocabulary from unlabeled images. As a result, expensive mask decoding is performed only for prompts supported by the image, while the final prediction uses the joint response across classes.

\subsection{Contextual Prompt Expansion}
\label{subsec:cpe}

Class names in open-vocabulary segmentation are often short category labels rather than natural text prompts. Compound labels such as \emph{windowpane} and \emph{trafficlight}, for example, may be harder for the text encoder to interpret than their natural forms, \emph{window pane} and \emph{traffic light}. To obtain clearer class prompts, our \emph{Contextual Prompt Expansion} (CPE) first canonicalizes each class name and then adds two complementary sources of context: related classes from the vocabulary and WordNet hypernyms. For each class $c\in\mathcal{C}$, a fixed lexical canonicalizer $\eta$ converts the original class name into a canonicalized form $\tilde{c}=\eta(c)$, such as \emph{windowpane}$\mapsto$\emph{window pane}. The frozen SAM~3 text encoder \citep{carion2025sam} then maps $\tilde{c}$ to its prompt tokens $Z_c$ and pooled text embedding $e_c$. Using $e_c$, we find the $M_s=2$ other classes in the vocabulary with the highest cosine similarity and project their pooled embeddings into prompt tokens, denoted by $T_c^{\mathrm{nbr}}$. We also obtain up to $M_t=2$ noun hypernyms of $\tilde{c}$ from WordNet \citep{miller1995wordnet} and encode them in the same way to obtain $T_c^{\mathrm{hyp}}$. The final contextual prompt is
\begin{equation}
\bar{Z}_c = Z_c \mathbin{\Vert} T_c^{\mathrm{nbr}} \mathbin{\Vert} T_c^{\mathrm{hyp}},
\label{eq:cpe-prompt}
\end{equation}
where $\bar{Z}_c$ is the contextual prompt used for class $c$, $Z_c$ contains the tokens of its canonicalized class name, $T_c^{\mathrm{nbr}}$ contains the $M_s=2$ vocabulary-neighbor tokens, and $T_c^{\mathrm{hyp}}$ contains up to $M_t=2$ WordNet hypernym tokens. The operator $\mathbin{\Vert}$ denotes concatenation along the token sequence. All three sources depend only on the class vocabulary and fixed lexical resources: the canonicalization rules and WordNet lookup are fixed before evaluation and use no images or annotations. The resulting prompts are constructed once for each vocabulary and cached for both full-resolution inference and the signature-estimation pass in Section~\ref{subsec:calibration}. The low-resolution presence preview in Section~\ref{subsec:epg} uses only the canonicalized class tokens $Z_c$, without the additional context tokens, to keep the preview inexpensive. Appendix~\ref{app:implementation} gives further details.

\subsection{Evidence-Proportional Grounding}
\label{subsec:epg}

Most images contain only a small subset of the classes in the full vocabulary. ActiveSAM therefore uses the SAM~3 presence head to remove unlikely classes before expensive full-resolution decoding. A single global presence threshold $\tau$ is used at both the preview and full-resolution stages.

\paragraph{Presence preview.}
We run frozen SAM~3 at a preview resolution of 336 over all $C$ class prompts, using the canonicalized class tokens $Z_c$ without the additional CPE context tokens. Since this stage only estimates class presence, it uses the image--text fusion encoder and a shortened object decoder while skipping the segmentation head. This shortened preview gives the same class-admission decisions as the full decoder on all tested benchmarks (Appendix~\ref{app:implementation}). Let $q_c^{\mathrm p}=\operatorname{sigmoid}(\rho_c)$ denote the preview presence score of class $c$, where $\rho_c$ is the corresponding SAM~3 presence logit. We admit
\begin{equation}
\mathcal{A}(I)=\{c\in\mathcal{C}: q_c^{\mathrm p}\ge\tau\}\cup\{\bg\},
\label{eq:active-set}
\end{equation}
where $\mathcal{A}(I)$ is the active set of image $I$, the classes admitted by the preview, and $\tau$ is the single global presence threshold shared across all datasets. For benchmarks with a background class, the background prompt $\bg$ is always admitted, and for other benchmarks, $\bg$ is omitted. If no semantic class reaches $\tau$, we retain the class with the highest presence score. Thus, only classes with sufficient presence evidence proceed to the full-resolution stage. Classes rejected by the preview require no full-resolution decoding, whereas SAM~3 \citep{carion2025sam} and SegEarth-OV3 \citep{li2025segearth} decode the full vocabulary before applying presence scores. We use a preview resolution of 336 throughout, and performance is stable across a wide range of preview resolutions (Table~\ref{tab:preview_resolution}).

\paragraph{Full-resolution decoding.}
ActiveSAM runs the full-resolution SAM~3 image encoder once at resolution 1008 and divides the active classes into buckets of at most $K=32$. Classes in the same bucket are processed together in one decoder forward pass. For each bucket, the image--text fusion encoder and the first decoder layer produce a full-resolution presence score $q_c^{(1)}$ for class $c$. If $q_c^{(1)}<\tau$, the class is stopped early and assigned an all-zero score map. Otherwise, it is processed by the full decoder, segmentation head, and post-processing. Following SegEarth-OV3 \citep{li2025segearth}, we obtain instance and semantic score maps and fuse them with the final presence score $q_c$,
\begin{equation}
S_c(p)=q_c\,\max\bigl(S_c^{\mathrm{inst}}(p),S_c^{\mathrm{sem}}(p)\bigr),
\label{eq:head-fusion}
\end{equation}
where $S_c(p)$ is the final score of class $c$ at pixel $p$, $S_c^{\mathrm{inst}}(p)$ and $S_c^{\mathrm{sem}}(p)$ are its instance and semantic scores, and $q_c$ is its final full-resolution presence score. Classes rejected before full decoding are assigned zero scores, i.e., $S_c(p)=0$ for all pixels $p$. We denote the full class-score vector at pixel $p$ by $S(p)\in[0,1]^C$, and the standard one-versus-rest prediction by $y_0(p)=\arg\max_c S_c(p)$.

\subsection{Exclusive Concept Decoding}
\label{subsec:ecd}

SAM~3 predicts a score map for each class prompt independently, so related classes can respond to the same pixels. For example, a \emph{rider} can also receive a high \emph{person} score, and an \emph{armchair} can also receive a high \emph{chair} score. A pixelwise argmax $y_0$ considers only the largest score and can therefore favor the broader class. Exclusive Concept Decoding (ECD) instead uses the joint score pattern across classes.

\noindent{{\textbf{Response signatures.}}}
For each class $k$, we define a response signature $R_k\in[0,1]^C$ as the expected class-score vector on pixels selected for that class by the frozen model. For example, a \emph{rider} signature can contain high responses for both \emph{rider} and \emph{person}, while a \emph{person} signature need not respond strongly to \emph{rider}. The signatures are estimated once per vocabulary from unlabeled images, together with the pixel share $m_k$ of each class (Section~\ref{subsec:calibration}).

\paragraph{Decoding.}
For each image, let $S_{\mathcal{A}}(p)$ denote the score vector $S(p)$ restricted to the active classes $\mathcal{A}(I)$, and let $R_{k,\mathcal{A}}$ denote the corresponding entries of class signature $R_k$. We consider classes with a positive score at pixel $p$, $\mathcal{K}(p)=\{k\in\mathcal{A}(I):S_k(p)>0\}$. ECD combines the class prior with its distance to the response signature and selects the class with the highest score
\begin{equation}
y(p)=\arg\max_{k\in\mathcal{K}(p)}\left[\log(q_km_k)-\frac{\|S_{\mathcal{A}}(p)-R_{k,\mathcal{A}}\|_2^2}{2\sigma^2}\right],
\label{eq:ecd}
\end{equation}
where $y(p)$ is the ECD prediction, $\|S_{\mathcal{A}}(p)-R_{k,\mathcal{A}}\|_2^2$ is the squared Euclidean distance between the pixel score vector and the class signature, $q_k$ is the final full-resolution presence score of class $k$, and $m_k$ is its pixel share on the shared unlabeled pool. The term $\log(q_km_k)$ accounts for support from the current image and the unlabeled pool, while the distance term measures how closely the pixel score pattern matches each class signature. The scale $\sigma$ controls the balance between these two terms and is shared across all datasets.

\paragraph{Agreement and support.}
We evaluate Eq.~\ref{eq:ecd} using the two signature estimates $R_k^{\mathrm a}$ and $R_k^{\mathrm e}$ defined in Section~\ref{subsec:calibration}. We accept the ECD prediction only when both signature estimates give the same class, otherwise the pixel keeps $y_0(p)$. Classes with insufficient support in the unlabeled pool also keep the baseline prediction.

\paragraph{Null-signature background.}
For benchmarks with an explicit background class, we additionally represent background by a null signature $R_{\bg}=\mathbf{0}$. Its prior is determined by the background mass $m_{\bg}$ estimated from the shared unlabeled pool and a single global offset $b$ shared across datasets. A pixel assigned to class $y$ is changed to background when the null signature receives the higher score,
\begin{equation}
\log m_{\bg}+b-\frac{\|S_{\mathcal{A}}(p)\|_2^2}{2\sigma^2}>\log(q_ym_y)-\frac{\|S_{\mathcal{A}}(p)-R^{\mathrm a}_{y,\mathcal{A}}\|_2^2}{2\sigma^2},
\label{eq:null-background}
\end{equation}
where $\|S_{\mathcal{A}}(p)\|_2^2$ is the squared Euclidean distance from the pixel score vector to the null background signature, while $\|S_{\mathcal{A}}(p)-R^{\mathrm a}_{y,\mathcal{A}}\|_2^2$ is its squared Euclidean distance to the signature of the predicted class $y$. The pixel is therefore assigned to background only when the null signature gives the higher score. This uses the same signature comparison as the semantic classes and avoids a dataset-specific background threshold.

\subsection{Label-free Signature Estimation}
\label{subsec:calibration}

We estimate the response signatures from a shared unlabeled subset of COCO train2017 images \citep{lin2014microsoft}. The same image pool is used across all benchmarks in this work, without using semantic annotations or any evaluation image. Each pool image is processed by CPE and EPG without ECD, producing the class-score maps used to estimate the signatures. This pass only aggregates outputs from the frozen model and does not update any model parameters.

For each class $k$, we collect three groups of pixels: pixels where $k$ has the highest score, pixels where $k$ responds strongly, and pixels where $k$ responds without a competing class. Let $\mu_k^{\mathrm a}$, $\mu_k^{\mathrm p}$, and $\mu_k^{\mathrm e}$ denote the mean class-score vectors over these three groups, respectively. The first ECD signature is $R_k^{\mathrm a}=\mu_k^{\mathrm a}$. The second signature $R_k^{\mathrm e}$ uses the exclusive estimate $\mu_k^{\mathrm e}$ when it has sufficient support, and otherwise falls back to $\mu_k^{\mathrm p}$ and then $\mu_k^{\mathrm a}$. We also estimate the class mass $m_k$ from the fraction of pixels where class $k$ has the highest score.

These two signatures capture complementary views of the same class. $R_k^{\mathrm a}$ reflects the response pattern where the class wins against other classes, while $R_k^{\mathrm e}$ aims to capture its response when competing classes are weak. Their agreement is used by ECD to avoid uncertain reassignments. All signatures and class masses are estimated once per vocabulary from the shared unlabeled pool. The exact pixel-selection thresholds, support rule, pool construction, and implementation details are given in Appendix~\ref{app:calibration}.

\section{Experimental Results}
\noindent\textbf{Experimental Setup.}
We evaluate ActiveSAM on eight standard OVSS benchmarks: Pascal VOC 20/21 (with a background class) \citep{everingham2010pascal}, Pascal Context 59/60 \citep{mottaghi2014role}, COCO-Object \citep{lin2014microsoft}, COCO-Stuff \citep{caesar2018coco}, Cityscapes \citep{cordts2016cityscapes}, and ADE20K \citep{zhou2017scene}. COCO-Object evaluates the 80 MS COCO classes and treats ``stuff'' as background. Following the standard practice in training-free OVSS, we use no target-dataset annotations such as ground-truth class presence or pixel-level labels, and no evaluation image is used before inference. All experiments are conducted on a single NVIDIA RTX 5090, and performance is reported using mean Intersection over Union (mIoU). We use the same hyperparameters across all datasets without per-benchmark tuning: $\tau=0.1$, $\sigma=0.15$, pixel-selection thresholds $0.5$ and $0.3$, minimum support $N=2000$, and $b=1$. The preview resolution is 336. Our hyperparameter analysis (Table~\ref{tab:constants}) and preview-resolution analysis (Table~\ref{tab:preview_resolution}) support these choices: each default is within 0.04 mean mIoU of the best tested value, and performance remains stable across all tested values. Further implementation details are provided in Appendix~\ref{app:implementation}.

\subsection{Main Results}
\noindent\textbf{Comparison with state-of-the-art methods.}
As summarized in Table~\ref{tab:main_results}, ActiveSAM achieves the highest mean mIoU of $66.0$ across eight OVSS benchmarks, outperforming the current state-of-the-art method, SegEarth-OV3 \citep{li2025segearth}, by $2.1$ points and SAM~3 \citep{carion2025sam} by $8.5$ points. The largest gains over SAM~3 are on COCO-Stuff ($+11.6$), Cityscapes ($+10.7$), and Pascal Context-60/59 ($+10.5$/$+10.4$). Compared with SegEarth-OV3, which uses the same SAM~3 backbone, ActiveSAM improves VOC21 by $3.6$ points, Cityscapes by $3.3$, Context60 by $3.2$, ADE20K by $2.4$, and COCO-Stuff by $2.1$. Together with the efficiency results in Table~\ref{tab:fps_vs_mIoU_per_dataset}, these gains support our motivation to avoid full-vocabulary decoding when most classes are absent from an image. Qualitative comparisons with SegEarth-OV3 are shown in Appendix~\ref{app:qualitative}.
\begin{table}[t]
    \centering
    \caption{\textbf{Open-vocabulary semantic segmentation on 8 standard benchmarks.} (mIoU, \%). The best score in each column is shown in \textbf{bold}. Further baselines are listed in Table~\ref{tab:additional_baselines} (Appendix~\ref{app:baselines}). DI and DI2 denote DINO and DINOv2, respectively. S/16, B/8, B/14, B/16, L/14, and L/16 denote ViT-S/16, ViT-B/8, ViT-B/14, ViT-B/16, ViT-L/14, and ViT-L/16, respectively. $\dagger$ indicates DINOv2 with registers. $\diamond$ = DI (B/8) + SAM2 (Hiera-L), $\ddagger$ = DI (B/16) + SAM (B/16), and $\S$ = DI2 (L/14) + SAM (L/16). P59/60: Pascal Context 59/60, C-O/C-S: COCO-Object/COCO-Stuff.}
    \label{tab:main_results}
    \resizebox{\linewidth}{!}{
    \begin{tabular}{@{}lllrccrrcrcc@{}}
    \toprule
    \textbf{Method} & \textbf{Venue} & \textbf{Backbone (Size)} & \multicolumn{3}{c}{\textbf{\textit{with background}}} & \multicolumn{5}{c}{\textbf{\textit{without background}}} & \multirow{2}{*}{\textbf{Avg.}} \\
    \cmidrule(lr){4-6} \cmidrule(lr){7-11}
    & & & \textbf{V21} & \textbf{P60} & \textbf{C-O} & \textbf{V20} & \textbf{P59} & \textbf{C-S} & \textbf{City} & \textbf{ADE} & \\
    \midrule
    \multicolumn{12}{l}{\textbf{Training-based}} \\
    GroupViT \citep{xu2022groupvit} & CVPR'22 & GroupViT (S/16) & 50.4 & 18.7 & 27.5 & 79.7 & 23.4 & 15.3 & 11.1 & 9.2 & 29.4 \\
    TCL \citep{cha2023learning} & CVPR'23 & CLIP & 51.2 & 24.3 & 30.4 & 77.5 & 30.3 & 19.6 & 23.1 & 14.9 & 33.9 \\
    CLIP-DINOiser \citep{wysoczanska2024clip} & ECCV'24 & CLIP + DI (B/16) & 62.1 & 32.4 & 34.8 & 80.9 & 35.9 & 24.6 & 31.7 & 20.0 & 40.3 \\
    Talk2DINO \citep{barsellotti2025talking} & ICCV'25 & CLIP + DI2$^{\dagger}$ (B/14) & 65.8 & 37.7 & 45.1 & 88.5 & 42.4 & 30.2 & 38.1 & 22.5 & 46.3 \\
    \midrule
    \multicolumn{12}{l}{\textbf{Training-free}} \\
    FreeDA \citep{barsellotti2024training} & CVPR'24 & CLIP + DI2 (B/14) & 51.8 & 35.3 & 36.3 & 84.3 & 39.7 & 25.7 & 34.1 & 20.8 & 41.0 \\
    ProxyCLIP \citep{lan2024proxyclip} & ECCV'24 & CLIP + DI2$^{\dagger}$ (B/14) & 58.6 & 33.8 & 37.4 & 83.0 & 37.2 & 25.4 & 33.9 & 19.7 & 41.1 \\
    CASS \citep{kim2025distilling} & CVPR'25 & CLIP + DI (B/8) & 65.8 & 36.7 & 37.8 & 87.8 & 40.2 & 26.7 & 39.4 & 20.4 & 44.4 \\
    CorrCLIP \citep{zhang2025corrclip} & ICCV'25 & CLIP + $\diamond$ & 76.7 & 44.9 & 49.4 & 91.5 & 50.8 & 34.0 & 51.1 & 30.7 & 53.6 \\
    Trident \citep{shi2025harnessing} & ICCV'25 & CLIP + $\ddagger$ & 67.1 & 38.6 & 41.1 & 84.5 & 42.2 & 28.3 & 42.9 & 21.9 & 45.8 \\
    ReME \citep{xuan2025reme} & ICCV'25 & CLIP + $\S$ & 82.2 & 44.6 & 48.2 & 93.2 & 53.1 & 33.3 & 59.0 & 28.2 & 55.2 \\
    SegEarth-OV3 \citep{li2025segearth} & arXiv'25 & SAM~3 (PE-L+/14) & 79.8 & 53.4 & 72.0 & 96.8 & 59.2 & 42.8 & 69.7 & 37.6 & 63.9 \\
    RF-CLIP \citep{li2026target} & AAAI'26 & CLIP & 67.2 & 37.9 & 39.1 & 87.0 & 41.4 & 27.5 & 43.0 & 21.0 & 45.5 \\
    SAM~3 \citep{carion2025sam} & ICLR'26 & SAM~3 (PE-L+/14) & 81.9 & 46.1 & 65.4 & 88.9 & 50.0 & 33.3 & 62.3 & 31.8 & 57.5 \\
    \textbf{ActiveSAM} & Ours & SAM~3 (PE-L+/14) & \textbf{83.4} & \textbf{56.6} & \textbf{72.6} & \textbf{97.1} & \textbf{60.4} & \textbf{44.9} & \textbf{73.0} & \textbf{40.0} & \textbf{66.0} \\
    \bottomrule
    \end{tabular}
    }
\end{table}

\begin{table}[t]
    \centering
    \caption{\textbf{Per-dataset mIoU and FPS comparison} on the eight OVSS benchmarks where ``$+\Delta$'' is the mIoU improvement. ``$N\times$'' means ActiveSAM is $N$ times faster than the respective method on the corresponding dataset. Best score per column is in \textbf{bold}.}
    \label{tab:fps_vs_mIoU_per_dataset}
    \renewcommand{\arraystretch}{1.15}
    \newcommand{\gpos}[1]{\textcolor{green!55!black}{\scriptsize\,(#1)}}
    \setlength{\tabcolsep}{2pt}
    \footnotesize
    \resizebox{\linewidth}{!}{
    \begin{tabular}{l l c c c c c c c c c}
    \toprule
    \textbf{Method} & \textbf{Metric} & \textbf{ADE20K} & \textbf{C59} & \textbf{C60} & \textbf{CO-O} & \textbf{CO-S} & \textbf{V21} & \textbf{V20} & \textbf{City} & \textbf{Mean} \\
    \midrule
    \multirow{2}{*}{CASS \citep{kim2025distilling}}
      & mIoU & 20.4\,\gpos{+19.6}       & 40.2\,\gpos{+20.2}       & 36.7\,\gpos{+19.9}       & 37.8\,\gpos{+34.8}       & 26.7\,\gpos{+18.2}       & 65.8\,\gpos{+17.6}        & 87.8\,\gpos{+9.3}         & 39.4\,\gpos{+33.6}        & 44.4\,\gpos{+21.6}       \\
      & FPS  & 0.70\,\gpos{$5.0\times$} & 0.81\,\gpos{$7.8\times$} & 0.81\,\gpos{$7.6\times$} & 0.68\,\gpos{$10.5\times$} & 0.63\,\gpos{$6.0\times$} & 2.30\,\gpos{$4.1\times$}  & 0.82\,\gpos{$11.6\times$} & 0.11\,\gpos{$36.2\times$} & 0.86\,\gpos{$11.1\times$} \\
    \cmidrule(lr){1-11}
    \multirow{2}{*}{SegEarth-OV3 \citep{li2025segearth}}
      & mIoU & 37.6\,\gpos{+2.4}        & 59.2\,\gpos{+1.2}        & 53.4\,\gpos{+3.2}        & 72.0\,\gpos{+0.6}        & 42.8\,\gpos{+2.1}        & 79.8\,\gpos{+3.6}         & 96.8\,\gpos{+0.3}         & 69.7\,\gpos{+3.3}         & 63.9\,\gpos{+2.1}        \\
      & FPS  & 0.35\,\gpos{$10.0\times$} & 0.86\,\gpos{$7.4\times$} & 0.84\,\gpos{$7.3\times$} & 0.63\,\gpos{$11.4\times$} & 0.31\,\gpos{$12.2\times$} & 2.19\,\gpos{$4.3\times$}  & 1.92\,\gpos{$5.0\times$}  & 1.81\,\gpos{$2.2\times$}  & 1.11\,\gpos{$7.5\times$} \\
    \cmidrule(lr){1-11}
    \multirow{2}{*}{SAM~3 \citep{carion2025sam}}
      & mIoU & 31.8\,\gpos{+8.2}         & 50.0\,\gpos{+10.4}        & 46.1\,\gpos{+10.5}        & 65.4\,\gpos{+7.2}         & 33.3\,\gpos{+11.6}        & 81.9\,\gpos{+1.5}         & 88.9\,\gpos{+8.2}         & 62.3\,\gpos{+10.7}        & 57.5\,\gpos{+8.5}        \\
      & FPS  & 0.13\,\gpos{$27.0\times$} & 0.32\,\gpos{$19.8\times$} & 0.32\,\gpos{$19.1\times$} & 0.24\,\gpos{$29.8\times$} & 0.11\,\gpos{$34.5\times$} & 0.93\,\gpos{$10.1\times$} & 0.93\,\gpos{$10.3\times$} & 0.76\,\gpos{$5.2\times$}  & 0.47\,\gpos{$19.5\times$} \\
    \cmidrule(lr){1-11}
    \multirow{2}{*}{\textbf{ActiveSAM}}
      & mIoU & \textbf{40.0} & \textbf{60.4} & \textbf{56.6} & \textbf{72.6} & \textbf{44.9} & \textbf{83.4} & \textbf{97.1} & \textbf{73.0} & \textbf{66.0} \\
      & FPS  & \textbf{3.51} & \textbf{6.33} & \textbf{6.12} & \textbf{7.16} & \textbf{3.79} & \textbf{9.37} & \textbf{9.54} & \textbf{3.98} & \textbf{6.23} \\
    \bottomrule
    \end{tabular}
    }
    \vspace{-0.15in}
    \end{table}
\noindent\textbf{Speed–accuracy advantage.}
ActiveSAM achieves a substantial efficiency gain across all eight benchmarks. As shown in Table~\ref{tab:fps_vs_mIoU_per_dataset}, it runs much faster than both SAM~3 \citep{carion2025sam} and SegEarth-OV3 \citep{li2025segearth} on every benchmark. Compared with the current state-of-the-art SegEarth-OV3, ActiveSAM runs $7.3$--$12.2\times$ faster on Context59/60, COCO-Object, COCO-Stuff, and ADE20K, while also improving mIoU on all five datasets. Across all eight benchmarks, ActiveSAM is $7.5\times$ faster on average. These results highlight a key strength of ActiveSAM: it achieves both goals at once, substantially faster inference and improved segmentation accuracy.

\noindent\textbf{Robustness under severe image corruptions.}
Practical segmentation systems should remain reliable under input degradation. Autonomous vehicles encounter fog and motion blur. We therefore evaluate robustness using four representative ImageNet-C corruptions \citep{hendrycks2019benchmarking}. All corruptions are evaluated at severity 5. ActiveSAM achieves the highest mean mIoU under all four corruptions, outperforming the current state-of-the-art SegEarth-OV3 \citep{li2025segearth} by $1.6$-$3.6$ points across six benchmarks (Table~\ref{tab:corruption_robustness}). These results show that ActiveSAM's accuracy advantage extends beyond clean images and remains consistent under severe input corruption.
\begin{table}[t]
\centering
\caption{\textbf{Prompt selection and bucketing.} We remove one component at a time to measure its contribution to the speedup. Without bucketing, each active prompt is decoded in its own decoder pass instead of in buckets, which leaves mIoU unchanged. Without prompt selection, every prompt is decoded with full decoding layers.}
\label{tab:bucketing}
\setlength{\tabcolsep}{7pt}
\footnotesize
\resizebox{0.78\linewidth}{!}{\begin{tabular}{@{}l cc cc@{}}
\toprule
\multirow{2}{*}{\textbf{Dataset}} & \multicolumn{2}{c}{\textbf{Slowdown vs.\ ActiveSAM}} & \multicolumn{2}{c}{\textbf{Speedup over SegEarth-OV3}} \\
\cmidrule(lr){2-3} \cmidrule(lr){4-5}
 & w/o bucketing & w/o prompt selection & ActiveSAM & w/o bucketing \\
\midrule
VOC21       & $1.28\times$ & $1.84\times$ & $4.3\times$ & $3.3\times$ \\
VOC20       & $1.22\times$ & $2.17\times$ & $5.0\times$ & $4.1\times$ \\
Cityscapes  & $1.74\times$ & $1.25\times$ & $2.2\times$ & $1.3\times$ \\
Context59   & $1.87\times$ & $2.65\times$ & $7.4\times$ & $3.9\times$ \\
Context60   & $1.92\times$ & $2.55\times$ & $7.3\times$ & $3.8\times$ \\
COCO-Object & $1.53\times$ & $3.96\times$ & $11.4\times$ & $7.4\times$ \\
COCO-Stuff  & $2.15\times$ & $4.12\times$ & $12.2\times$ & $5.7\times$ \\
ADE20K      & $2.25\times$ & $3.34\times$ & $10.0\times$ & $4.5\times$ \\
\midrule
Mean        & $1.74\times$ & $2.73\times$ & $7.5\times$ & $4.3\times$ \\
\bottomrule
\end{tabular}}
\end{table}
\subsection{Ablation studies and Analysis}
\noindent\textbf{Component ablations.}
We ablate each component separately: CPE and ECD in Table~\ref{tab:component_ablation}, and EPG in Table~\ref{tab:epg_ablation}. CPE and ECD mainly improve accuracy, whereas EPG mainly reduces computation. Removing CPE lowers mean mIoU by 1.4 points, with the largest drop on VOC21 (83.4 to 79.0). Among the CPE components, lexical canonicalization contributes the most (Table~\ref{tab:cpe_parts}), showing that clear class names matter for concept prompting. Replacing ECD with a pixelwise argmax lowers mean mIoU by 0.7 points. The gain is concentrated on Cityscapes ($+3.3$), where \emph{rider} overlaps \emph{person}, and on Context60 ($+1.5$). On the remaining benchmarks, ECD changes mIoU by at most 0.6 points. With $\tau=0.1$, the preview keeps on average only 7.3 of 81 prompts per image on COCO-Object and 30.8 of 150 on ADE20K (Table~\ref{tab:epg_retention}). Decoding only these prompts changes mIoU by at most 0.34 on any benchmark, yet makes inference 1.8--4.1$\times$ faster on seven of the eight benchmarks (Table~\ref{tab:epg_ablation}). Cityscapes is the exception ($1.25\times$): its images contain most of its classes, so the preview keeps 18.0 of 23 prompts and there is little to skip.

\begin{table}[t]
\centering
    \caption{\textbf{Component ablation.} mIoU (\%) after removing ECD or CPE. The EPG ablation is reported in Table~\ref{tab:epg_ablation}.}
\label{tab:component_ablation}
\renewcommand{\arraystretch}{1.15}
\resizebox{\linewidth}{!}{
\begin{tabular}{@{}l ccc ccccc cc@{}}
\toprule
\multirow{2}{*}{\textbf{Configuration}} & \multicolumn{3}{c}{\textbf{\textit{with background}}} & \multicolumn{5}{c}{\textbf{\textit{without background}}} & \multirow{2}{*}{\textbf{Mean}} & \multirow{2}{*}{\textbf{$\Delta$ vs Full}} \\
\cmidrule(lr){2-4} \cmidrule(lr){5-9}
 & \textbf{V21} & \textbf{P60} & \textbf{C-O} & \textbf{V20} & \textbf{P59} & \textbf{C-S} & \textbf{City} & \textbf{ADE} & & \\
\midrule
\textbf{Full ActiveSAM (CPE + EPG + ECD)} & 83.4 & 56.6 & 72.6 & 97.1 & 60.4 & 44.9 & 73.0 & 40.0 & \textbf{66.0} & --- \\
\quad $-$ ECD \,(pixelwise argmax) & 83.3 & 55.1 & 72.2 & 97.1 & 60.5 & 44.3 & 69.7 & 40.0 & 65.3 & $-0.7$ \\
\quad $-$ CPE \,(plain prompts, signatures re-estimated) & 79.0 & 56.1 & 72.8 & 96.7 & 60.3 & 43.4 & 70.1 & 38.8 & 64.6 & $-1.4$ \\
\midrule
SegEarth-OV3 \citep{li2025segearth} (for reference) & 79.8 & 53.4 & 72.0 & 96.8 & 59.2 & 42.8 & 69.7 & 37.6 & 63.9 & $-2.1$ \\
\bottomrule
\end{tabular}
}
\end{table}
\noindent\textbf{Prompt selection and bucketing.}
Table~\ref{tab:bucketing} removes prompt selection and bucketing one at a time. Prompt selection provides the largest share of the speedup: decoding every prompt slows ActiveSAM by $2.73\times$ on average and by $3.3$--$4.1\times$ on COCO-Object, COCO-Stuff and ADE20K. Bucketed decoding also contributes substantially: removing it slows ActiveSAM by $1.74\times$ on average without changing mIoU. Even without bucketing, ActiveSAM remains $4.3\times$ faster than SegEarth-OV3 \citep{li2025segearth} on average, keeping most of its $7.5\times$ faster inference.

\noindent\textbf{Shared calibration pool.}
ActiveSAM does not select calibration images per benchmark: the signatures for all eight vocabularies are estimated from the same pool of unlabeled COCO training images, and no evaluation image is used before inference. To measure what this stricter setting costs, we also estimate the signatures from each benchmark's own evaluation images, without labels (Table~\ref{tab:calibration_source}). This variant raises mean mIoU only slightly, from 66.02 to 66.40, so the shared pool remains competitive with calibrating on each benchmark's own images. We keep the shared pool for all other results because it follows the stricter protocol.

\noindent\textbf{One configuration for all benchmarks.}
ActiveSAM uses the same hyperparameters on all eight benchmarks, whereas SegEarth-OV3 \citep{li2025segearth} adjusts its confidence and background thresholds per dataset. A shared configuration matters in open-vocabulary use, where a new vocabulary rarely comes with labeled data for tuning. Table~\ref{tab:constants} tests whether this configuration depends on its exact values: it varies six hyperparameters one at a time and disables the agreement rule and the class-mass term. All 14 variants keep mean mIoU at 65.1 or higher, above SegEarth-OV3's 63.9, and 13 of them stay within 0.5 points of the default. The only larger drop, 0.9 points, comes from raising the signature-distance scale $\sigma$ to 0.20, which gives more weight to the class prior.

\section{Conclusion}
\label{sec:conclusion}
We presented ActiveSAM, a training-free framework for open-vocabulary semantic segmentation with frozen SAM~3. CPE clarifies class prompts, EPG selects which prompts receive full-resolution mask prediction, and ECD uses joint class responses to distinguish overlapping concepts. The response signatures are estimated once per vocabulary from a shared unlabeled image pool. Across eight benchmarks, ActiveSAM improves mean mIoU by 2.1 over SegEarth-OV3 \citep{li2025segearth} and runs much faster on every benchmark, by $7.3\times$ to $12.2\times$ on the large-vocabulary benchmarks. It also achieves the highest segmentation accuracy under severe image corruptions, which is desirable in real-world conditions. These results show that selecting relevant prompts and resolving their overlap can improve both the efficiency and accuracy of a frozen segmentation model.

\subsection*{AI use statement}
In this work, we used generative AI tools to polish writing and improve grammar. We also used them to refine the format of tables and figures. All research ideas, methods, experiments, analyses, and substantive writing were carried out by the authors without AI/LLM assistance.

\subsection*{Ethics statement}
ActiveSAM may reduce the computational cost of deploying segmentation models, but it could also be misused for surveillance or tracking. Its English lexical resources and calibration pool may underrepresent some concepts. Prompt filtering can also omit small or occluded objects. We use public benchmark datasets under their respective licenses and collected no new personal data. These limitations require evaluation in the intended domain before any safety-critical use.

\subsection*{Reproducibility statement}
We have made efforts to ensure the reproducibility of our work. Section~\ref{sec:method} and Appendix~\ref{app:implementation} describe the method and detailed implementation settings. Appendix~\ref{app:calibration} details the signature estimation. Appendices~\ref{app:calibration_source} and~\ref{app:constants} examine calibration sources and hyperparameter sensitivity. We commit to providing our source code of implementation upon publication to facilitate reproduction of our results.

\bibliographystyle{iclr2027_conference}
\bibliography{iclr2027_conference}

\newpage
\appendix

\section*{\Large{Appendix}}
\section{Implementation Details}
\label{app:implementation}
We use the official SAM~3 model \citep{carion2025sam} with the Perception Encoder-Large+ backbone \citep{bolya2025perception} and keep all weights frozen. Images are resized to $1008\times1008$ for full-resolution inference and $336\times336$ for the preview. Together with the bucket size $K=32$, $M_s=M_t=2$ context tokens and the instance confidence threshold of 0.3, these settings are shared by all eight benchmarks. Class names pass through the fixed canonicalizer $\eta$ before text encoding. We use no test-time augmentation or additional mask refinement. For the speed--accuracy comparison, all methods are run from their official implementations on one NVIDIA RTX 5090 under the same evaluation setup.

\paragraph{Contextual prompt expansion.}
CPE adds two vocabulary-neighbor tokens and up to two WordNet hypernym tokens to each canonicalized class prompt. Each added token is the pooled text embedding of the neighbor class or hypernym, mapped by a frozen projection from the 1024-dimensional pooled space to the 256-dimensional prompt-token space. For WordNet, we take the first noun synset of the canonicalized name and its first two immediate hypernyms; if none exist, no hypernym tokens are added. The canonicalizer $\eta$ is a fixed set of string rules that rewrite class names into clearer forms, such as splitting compound words; it is applied identically to all datasets. The contextual prompts $\bar{Z}_c$ are cached once per vocabulary, so prompt construction is not repeated for each image.

\paragraph{Evidence-proportional grounding.}
The preview uses the canonicalized class tokens $Z_c$, the first two of SAM~3's six decoder layers, and the first 50 of its 200 object queries. It skips the segmentation head. On all tested benchmarks, this shortened preview admits the same classes as the full preview decoder. Prompts with presence at least $\tau=0.1$ are retained, along with the explicit background prompt where applicable. If no semantic prompt passes, the highest-scoring one is retained.

At full resolution, the active prompts are processed in vocabulary order in buckets of at most $K=32$, reusing the image features. After one decoder layer, prompts with presence below $\tau$ stop and receive zero score maps while retaining their one-layer presence scores. The remaining prompts use the full six-layer decoder, segmentation head, and post-processing. When a class has several names, each name is processed separately and the class score map and presence are the maxima over its names. This explains why VOC20 uses 26 prompts for 20 classes and Cityscapes uses 23 for 19.

\paragraph{Dual-head fusion and presence gating.}
Following SegEarth-OV3 \citep{li2025segearth}, we score each instance mask by the product of its class probability and prompt-presence score. Masks below the shared confidence threshold of 0.3 are discarded, and a pixelwise maximum over the remaining masks gives $S_c^{\mathrm{inst}}$. We fuse this map with the semantic map $S_c^{\mathrm{sem}}$ and multiply by the final full-resolution presence $q_c$, as in Eq.~\ref{eq:head-fusion}. Logits are upsampled to the original image size in float32 before applying the sigmoid. The final presence $q_c$ weights the score map, while preview presence $q_c^{\mathrm p}$ and first-layer presence $q_c^{(1)}$ control prompt selection.

\paragraph{Exclusive concept decoding.}
ECD loads the two signature matrices $R^{\mathrm a}$ and $R^{\mathrm e}$, class masses, background mass where applicable, and the list of classes with insufficient calibration support. It restricts score vectors and signatures to the active classes and evaluates Eq.~\ref{eq:ecd} with both signature estimates. A pixel keeps the baseline label $y_0(p)$ if the estimates disagree, either the original or proposed class lacks support, or no active class has a positive score. For background benchmarks, Eq.~\ref{eq:null-background} then uses $R^{\mathrm a}$ on pixels not protected by the support rule.

\section{Signature Estimation Details}
\label{app:calibration}
\paragraph{Calibration images.}
The shared pool consists of 9,858 COCO train2017 images \citep{lin2014microsoft}, evenly spaced in the file list and selected without regard to image content. Each image is processed with CPE and EPG, without ECD. The presence-weighted score maps are average-pooled to one quarter of the original height and width. No annotation file is used.

\paragraph{Pixel groups and signatures.}
For each class $k$, we collect pixels where it has the highest positive score, where its score is high, and where its score is high while competing scores are low:
\begin{equation}
\begin{aligned}
\Omega_k^{\mathrm a}&=\{p:\argmax\nolimits_j S_j(p)=k,\ \max\nolimits_j S_j(p)>0\},\\
\Omega_k^{\mathrm p}&=\{p:S_k(p)>0.5\},\\
\Omega_k^{\mathrm e}&=\{p:S_k(p)>0.5,\ \max\nolimits_{j\ne k}S_j(p)<0.3\}.
\end{aligned}
\label{eq:cores}
\end{equation}
Here, $p$ indexes pooled pixels across the calibration images, and $S_j(p)$ is the score of class $j$. Let $\mu_k^x$ be the mean score vector over $\Omega_k^x$ and $n_k^x=|\Omega_k^x|$ for $x\in\{\mathrm a,\mathrm p,\mathrm e\}$. We define the two signatures and class mass as
\begin{equation}
\begin{aligned}
R_k^{\mathrm a}&=\mu_k^{\mathrm a},\qquad m_k=\frac{n_k^{\mathrm a}}{\sum_j n_j^{\mathrm a}},\\
R_k^{\mathrm e}&=\begin{cases}
\mu_k^{\mathrm e} & n_k^{\mathrm e}\ge N,\\
\mu_k^{\mathrm p} & n_k^{\mathrm e}<N\le n_k^{\mathrm p},\\
\mu_k^{\mathrm a} & \text{otherwise}.
\end{cases}
\end{aligned}
\label{eq:signatures}
\end{equation}
The minimum support is $N=2000$ pooled pixels. A class with $n_k^{\mathrm a}<N$ is marked as uncalibrated, and ECD preserves the baseline label when either the original or proposed class is uncalibrated.

\section{Calibration Source}
\label{app:calibration_source}
Table~\ref{tab:calibration_source} changes only the unlabeled images used to estimate the signatures. Besides the default pool of about 10k images (Appendix~\ref{app:calibration}), we test a smaller 2k-image pool, which changes mean mIoU by only 0.02. For reference, we also estimate the signatures from each benchmark's own evaluation images, without labels. This variant raises mean mIoU from 66.02 to 66.40, but we do not use it, because our protocol never accesses evaluation images before inference. Apart from this reference variant, every experiment in the paper uses a shared COCO pool, whose train2017 images are disjoint from the val2017 images used to evaluate COCO-Object and COCO-Stuff.

\begin{table}[tbp]
\centering
\caption{\textbf{Calibration source.} mIoU (\%) with signatures estimated from different unlabeled image sets. The final row uses evaluation images without labels and is shown for reference only.}
\label{tab:calibration_source}
\setlength{\tabcolsep}{4pt}
\footnotesize
\resizebox{\linewidth}{!}{\begin{tabular}{@{}lrrrrrrrrr@{}}
\toprule
\textbf{Unlabeled pool} & \textbf{V21} & \textbf{P60} & \textbf{C-O} & \textbf{V20} & \textbf{P59} & \textbf{C-S} & \textbf{City} & \textbf{ADE} & \textbf{Mean} \\
\midrule
Shared COCO pool, 10k images (default) & 83.42 & 56.59 & 72.63 & 97.08 & 60.44 & 44.90 & 73.04 & 40.03 & 66.02 \\
Shared COCO pool, 2k images & 83.49 & 56.66 & 72.58 & 97.07 & 60.57 & 44.74 & 72.86 & 40.02 & 66.00 \\
\midrule
Evaluation images, no labels  & 83.63 & 56.52 & 72.75 & 97.10 & 60.43 & 44.83 & 74.89 & 41.02 & 66.40 \\
\bottomrule
\end{tabular}}
\end{table}

\section{Hyperparameter Sensitivity}
\label{app:constants}
Table~\ref{tab:constants} varies one setting at a time. Changing the presence threshold $\tau$, the two pixel-selection thresholds, or the minimum support $N$ changes mean mIoU by at most 0.42, and changing the background offset $b$ changes the mean over the three background benchmarks by at most 0.04. For the signature-distance scale $\sigma$, 0.10 and 0.15 give similar accuracy, while 0.20 gives more weight to the class prior and lowers mean mIoU to 65.13. Disabling the agreement rule or the class-mass term lowers mean mIoU by 0.22 and 0.27, respectively. Every variant still outperforms SegEarth-OV3 (63.9 mean mIoU, or 68.4 on the three background benchmarks), so the gains of ActiveSAM are stable and a single configuration can serve all eight benchmarks.

\begin{table}[tbp]
\centering
\caption{\textbf{Hyperparameter sensitivity.} One setting is varied at a time. Values are mean mIoU (\%) across eight benchmarks, except for $b$, which only affects the three benchmarks with a background class and is averaged over them (VOC21, Context60 and COCO-Object). The default setting is the middle value, or ``on'' for the two rules.}
\label{tab:constants}
\setlength{\tabcolsep}{5pt}
\footnotesize
\resizebox{\linewidth}{!}{\begin{tabular}{@{}llll@{}}
\toprule
\textbf{Setting} & \textbf{Role} & \textbf{Values tested} & \textbf{mIoU result} \\
\midrule
$\tau$ & prompt selection & 0.05 / 0.1 / 0.2 & 66.03 / 66.02 / 65.81 \\
$\sigma$ & signature distance versus prior & 0.10 / 0.15 / 0.20 & 65.92 / 66.02 / 65.13 \\
agreement rule & decode with $R^{\mathrm a}$ and $R^{\mathrm e}$ & on / off & 66.02 / 65.80 \\
mass term & prior $q_k m_k$ versus $q_k$ & on / off & 66.02 / 65.75 \\
confident-score threshold & confident and exclusive pixel groups & 0.4 / 0.5 / 0.6 & 65.62 / 66.02 / 66.05 \\
competitor-score threshold & competing scores in the exclusive group & 0.2 / 0.3 / 0.4 & 66.05 / 66.02 / 65.96 \\
$N$ & minimum support (pooled pixels) & 1000 / 2000 / 4000 & 65.60 / 66.02 / 66.03 \\
$b$ & background offset (three background benchmarks) & 0.5 / 1 / 1.5 & 70.86 / 70.88 / 70.92 \\
\bottomrule
\end{tabular}}
\end{table}

\section{CPE Components}
\label{app:cpe_parts}
Table~\ref{tab:cpe_parts} removes one CPE component at a time on the 2k pool and re-estimates the signatures with the resulting prompts. Lexical canonicalization contributes the most: removing it lowers mean mIoU by 1.06, with the largest drop on VOC21. Removing the neighbor or hypernym tokens lowers mean mIoU by 0.41 and 0.45, respectively. On the default 10k pool, replacing full CPE with plain prompts lowers mean mIoU by 1.37.

\begin{table}[tbp]
\centering
\caption{\textbf{CPE components.} mIoU (\%) after removing each prompt component and re-estimating signatures. Each $\Delta$ is relative to full CPE using the same pool. The upper block uses the 2k calibration pool, and the lower block uses the default 10k pool.}
\label{tab:cpe_parts}
\setlength{\tabcolsep}{4pt}
\footnotesize
\resizebox{\linewidth}{!}{\begin{tabular}{@{}lrrrrrrrrrr@{}}
\toprule
\textbf{Prompts} & \textbf{V21} & \textbf{P60} & \textbf{C-O} & \textbf{V20} & \textbf{P59} & \textbf{C-S} & \textbf{City} & \textbf{ADE} & \textbf{Mean} & \textbf{$\Delta$} \\
\midrule
Full CPE (2k pool) & 83.49 & 56.66 & 72.58 & 97.07 & 60.57 & 44.74 & 72.86 & 40.02 & 66.00 & --- \\
\quad without lexical canonicalization & 79.07 & 55.94 & 72.50 & 96.69 & 60.02 & 43.40 & 72.88 & 39.01 & 64.94 & $-1.06$ \\
\quad without neighbor tokens & 83.02 & 56.78 & 72.80 & 97.10 & 60.76 & 44.58 & 70.19 & 39.52 & 65.59 & $-0.41$ \\
\quad without hypernym tokens & 83.27 & 56.72 & 72.66 & 97.08 & 60.64 & 44.66 & 69.57 & 39.75 & 65.54 & $-0.45$ \\
\midrule
Full CPE (10k pool, default) & 83.42 & 56.59 & 72.63 & 97.08 & 60.44 & 44.90 & 73.04 & 40.03 & 66.02 & --- \\
\quad without CPE (plain prompts) & 79.02 & 56.07 & 72.84 & 96.70 & 60.25 & 43.40 & 70.09 & 38.82 & 64.65 & $-1.37$ \\
\bottomrule
\end{tabular}}
\end{table}

\section{Additional Baselines}
\label{app:baselines}
Table~\ref{tab:additional_baselines} reports further baselines under the protocol of Table~\ref{tab:main_results}. A dash marks an unreported result, and means are given only when all eight results are available. Among the baselines with all eight results, the strongest is SFP \citep{jin2025feature} at 44.0 mean mIoU, compared with 66.0 for ActiveSAM.
\begin{table}[tbp]
    \centering
    \caption{\textbf{Additional baselines.} mIoU (\%) on eight OVSS benchmarks.}
    \label{tab:additional_baselines}
    \resizebox{\linewidth}{!}{
    \begin{tabular}{@{}lllrccrrcrcc@{}}
    \toprule
    \textbf{Method} & \textbf{Venue} & \textbf{Backbone (Size)} & \textbf{V21} & \textbf{P60} & \textbf{C-O} & \textbf{V20} & \textbf{P59} & \textbf{C-S} & \textbf{City} & \textbf{ADE} & \textbf{Mean} \\
    \midrule
    \multicolumn{12}{l}{\textbf{Training-based}} \\
    SegCLIP \citep{luo2023segclip} & ICML'23 & CLIP & 52.6 & 24.7 & 27.5 & - & - & - & - & - & - \\
    CoDe \citep{wu2024image} & CVPR'24 & CLIP & 57.7 & 30.5 & 32.3 & - & - & 23.9 & 28.9 & 17.7 & - \\
    SAM-CLIP \citep{wang2024sam} & CVPR'24 & CLIP + SAM (ViT-B/16) & 60.6 & 29.2 & - & - & - & 31.5 & - & 17.1 & - \\
    \midrule
    \multicolumn{12}{l}{\textbf{Training-free}} \\
    CLIP \citep{radford2021learning} & ICML'21 & CLIP & 18.6 & 7.8 & 6.5 & 49.1 & 11.2 & 7.2 & 6.7 & 3.2 & 13.8 \\
    GEM \citep{bousselham2024grounding} & CVPR'24 & CLIP & 46.2 & - & - & - & 32.6 & 15.7 & - & - & - \\
    CaR \citep{sun2024clip} & CVPR'24 & CLIP & 48.6 & 13.6 & 15.4 & 73.7 & 18.4 & - & - & 5.4 & - \\
    LaVG \citep{kang2024defense} & ECCV'24 & CLIP + DINO (ViT-B/8) & 62.1 & 31.6 & 34.2 & 82.5 & 34.7 & 23.2 & 26.2 & 15.8 & 38.8 \\
    CLIPtrase \citep{shao2024explore} & ECCV'24 & CLIP & 50.9 & 29.9 & 43.6 & 81.0 & 33.8 & 22.8 & 21.3 & 16.4 & 37.5 \\
    ClearCLIP \citep{lan2024clearclip} & ECCV'24 & CLIP & 51.8 & 32.6 & 33.0 & 80.9 & 35.9 & 23.9 & 30.0 & 16.7 & 38.1 \\
    SCLIP \citep{wang2024sclip} & ECCV'24 & CLIP & 59.1 & 30.4 & 30.5 & 80.4 & 34.1 & 22.4 & 32.2 & 16.1 & 38.2 \\
    NACLIP \citep{hajimiri2025pay} & WACV'25 & CLIP & 58.9 & 32.2 & 33.2 & 79.7 & 35.2 & 23.3 & 35.5 & 17.4 & 39.4 \\
    LPOSS \citep{stojnic2025lposs} & CVPR'25 & CLIP + DINO (ViT-B/16) & 61.1 & 34.6 & 33.4 & 78.8 & 37.8 & 25.9 & 37.3 & 21.8 & 41.3 \\
    \textit{DIH}-CLIP \citep{duan2025dih} & ICCV'25 & CLIP & 64.2 & 36.0 & 37.4 & 84.9 & 39.7 & 26.7 & 40.2 & 19.6 & 43.6 \\
    SFP \citep{jin2025feature} & ICCV'25 & CLIP & 63.9 & 37.2 & 37.9 & 84.5 & 39.9 & 26.4 & 41.1 & 20.8 & 44.0 \\
    \midrule
    \textbf{ActiveSAM} & Ours & SAM~3 (PE-L+/14) & \textbf{83.4} & \textbf{56.6} & \textbf{72.6} & \textbf{97.1} & \textbf{60.4} & \textbf{44.9} & \textbf{73.0} & \textbf{40.0} & \textbf{66.0} \\
    \bottomrule
    \end{tabular}
    }
\end{table}

\section{Robustness to Image Corruptions}
\label{app:corruption}
Table~\ref{tab:corruption_robustness} reports mIoU under four ImageNet-C corruptions \citep{hendrycks2019benchmarking} at severity 5. ActiveSAM has the highest mean mIoU under every corruption and leads in 23 of the 24 dataset--corruption pairs; the exception is motion blur on Context60, where it reaches 42.5 mIoU versus 43.1 for SegEarth-OV3. Both CLIP-based methods stay far below the two SAM~3-based methods under every corruption, and under Gaussian noise their mean mIoU falls below 20.
\begin{table}[tbp]
    \centering
    \caption{\textbf{Robustness to image corruptions.} mIoU (\%) on clean inputs and four corruptions at severity 5. Means are over the six benchmarks shown. The best result for each condition and benchmark is bold. P60 and C-S denote Context60 and COCO-Stuff.}
    \label{tab:corruption_robustness}
    \resizebox{0.92\linewidth}{!}{
    \begin{tabular}{llccccccc}
        \toprule
        \textbf{Condition} & \textbf{Method} & \textbf{V20} & \textbf{V21} & \textbf{City} & \textbf{ADE} & \textbf{P60} & \textbf{C-S} & \textbf{Mean} \\
        \midrule
        \multirow{4}{*}{Clean}
            & CorrCLIP \citep{zhang2025corrclip}        & 91.5           & 76.7           & 51.1           & 30.7           & 44.9           & 34.0           & 54.8  \\
            & SegEarth-OV3 \citep{li2025segearth}     & 96.83          & 79.77          & 69.69          & 37.55          & 53.43          & 42.76          & 63.34 \\
            & RF-CLIP \citep{li2026target}         & 87.0           & 67.2           & 43.0           & 21.0           & 37.9           & 27.5           & 47.3  \\
            & ActiveSAM (Ours) & \textbf{97.08} & \textbf{83.42} & \textbf{73.04} & \textbf{40.03} & \textbf{56.59} & \textbf{44.90} & \textbf{65.84} \\
        \cmidrule(lr){1-9}
        \multirow{4}{*}{Gaussian noise}
            & CorrCLIP \citep{zhang2025corrclip}        & 35.66          & 25.25          &  8.99          &  5.51          & 11.55          &  9.15          & 16.02 \\
            & SegEarth-OV3 \citep{li2025segearth}    & 77.40          & 60.25          & 47.15          & 22.44          & 34.94          & 27.34          & 44.92 \\
            & RF-CLIP \citep{li2026target}         & 42.47          & 27.39          &  8.86          &  8.56          & 11.87          & 10.85          & 18.33 \\
            & ActiveSAM (Ours) & \textbf{84.22} & \textbf{66.21} & \textbf{47.72} & \textbf{24.32} & \textbf{38.29} & \textbf{30.27} & \textbf{48.51} \\
        \cmidrule(lr){1-9}
        \multirow{4}{*}{Motion blur}
            & CorrCLIP \citep{zhang2025corrclip}        & 62.64          & 47.91          & 37.04          & 15.69          & 28.39          & 19.62          & 35.21 \\
            & SegEarth-OV3 \citep{li2025segearth}    & 88.38          & 69.49          & 55.67          & 28.07          & \textbf{43.12} & 32.53          & 52.88 \\
            & RF-CLIP \citep{li2026target}         & 69.21          & 50.81          & 28.28          & 14.37          & 25.87          & 18.52          & 34.51 \\
            & ActiveSAM (Ours) & \textbf{90.62} & \textbf{70.00} & \textbf{59.21} & \textbf{30.15} & 42.50 & \textbf{34.40} & \textbf{54.48} \\
        \cmidrule(lr){1-9}
        \multirow{4}{*}{JPEG compression}
            & CorrCLIP \citep{zhang2025corrclip}        & 78.50          & 62.92          & 38.68          & 19.53          & 36.27          & 25.52          & 43.57 \\
            & SegEarth-OV3 \citep{li2025segearth}    & 92.30          & 74.96          & 58.40          & 31.55          & 45.10          & 35.46          & 56.29 \\
            & RF-CLIP \citep{li2026target}         & 71.71          & 52.39          & 26.21          & 15.22          & 26.70          & 19.19          & 35.24 \\
            & ActiveSAM (Ours) & \textbf{92.48} & \textbf{77.13} & \textbf{60.76} & \textbf{33.44} & \textbf{47.52} & \textbf{37.82} & \textbf{58.19} \\
        \cmidrule(lr){1-9}
        \multirow{4}{*}{Fog}
            & CorrCLIP \citep{zhang2025corrclip}        & 77.66          & 62.13          & 34.33          & 22.17          & 33.99          & 25.48          & 42.63 \\
            & SegEarth-OV3 \citep{li2025segearth}    & 94.74          & 79.44          & 64.18          & 34.98          & 50.15          & 40.63          & 60.69 \\
            & RF-CLIP \citep{li2026target}         & 76.55          & 53.43          & 25.66          & 16.55          & 29.82          & 22.15          & 37.36 \\
            & ActiveSAM (Ours) & \textbf{95.02} & \textbf{81.94} & \textbf{67.37} & \textbf{37.47} & \textbf{52.29} & \textbf{41.70} & \textbf{62.63} \\
        \bottomrule
    \end{tabular}
    }
\end{table}

\section{Ablation of Evidence-Proportional Grounding}
\label{app:epg}
Table~\ref{tab:epg_ablation} compares ActiveSAM with decoding every prompt at full resolution. EPG changes mIoU by at most 0.34 on any benchmark and by $-0.01$ on average, while making inference 1.8--4.1$\times$ faster on seven benchmarks. Cityscapes gains only $1.25\times$ because most of its prompts are admitted. Averaged over the eight benchmarks, EPG reduces the time per image from 0.521 to 0.184 seconds, and most of this saving comes from the preview.

\begin{table}[tbp]
\centering
\caption{\textbf{EPG ablation.} Full ActiveSAM (w/ preview) versus decoding every prompt at full resolution (w/o preview), on the full validation sets. FPS is end-to-end throughput on one RTX 5090 GPU, and speedup is the ratio of the two FPS values.}
\label{tab:epg_ablation}
\renewcommand{\arraystretch}{1.1}
\setlength{\tabcolsep}{4pt}
\footnotesize
\begin{tabular}{l r cc cc c}
\toprule
\multirow{2}{*}{\textbf{Dataset}} & \multirow{2}{*}{\textbf{Images}} & \multicolumn{2}{c}{\textbf{mIoU}} & \multicolumn{2}{c}{\textbf{FPS}} & \multirow{2}{*}{\textbf{Speedup}} \\
\cmidrule(lr){3-4} \cmidrule(lr){5-6}
 & & w/ preview & w/o preview & w/ preview & w/o preview & \\
\midrule
VOC21        & 1,449 & 83.42 & 83.35 & 9.37 & 5.09 & $1.84\times$ \\
VOC20        & 1,449 & 97.08 & 97.05 & 9.54 & 4.40 & $2.17\times$ \\
Cityscapes   &   500 & 73.04 & 72.99 & 3.98 & 3.19 & $1.25\times$ \\
Context59    & 5,105 & 60.44 & 60.51 & 6.33 & 2.39 & $2.65\times$ \\
Context60    & 5,105 & 56.59 & 56.50 & 6.12 & 2.40 & $2.55\times$ \\
COCO-Object  & 5,000 & 72.63 & 72.55 & 7.16 & 1.81 & $3.96\times$ \\
COCO-Stuff   & 5,000 & 44.90 & 45.24 & 3.79 & 0.92 & $4.12\times$ \\
ADE20K       & 2,000 & 40.03 & 40.05 & 3.51 & 1.05 & $3.34\times$ \\
\bottomrule
\end{tabular}
\end{table}

\paragraph{Class availability after filtering.}
Table~\ref{tab:epg_retention} reports the number of prompts admitted by the preview and the percentage of ground-truth image--class pairs that remain available after both EPG stages, that is, pairs whose class can still be predicted. Across the eight benchmarks, 93.8\% of these pairs remain available; the fraction is 93.9\% for connected components smaller than 1,024 pixels and 90.6\% for pairs in the least frequent third of each vocabulary. Retention is lowest on COCO-Stuff (79.8\%), yet decoding every prompt improves its mIoU by only 0.34, which suggests that many of the filtered classes were already difficult for the full pipeline.

\begin{table}[tbp]
\centering
\caption{\textbf{Prompt admission and class retention.} Mean active and total prompt counts use the preview threshold $\tau=0.1$. Retention is the percentage of ground-truth image--class pairs that remain available after both EPG stages. The total count prompts, and a class with several names contributes one prompt per name (VOC20: 26 prompts for 20 classes, Cityscapes: 23 for 19).}
\label{tab:epg_retention}
\setlength{\tabcolsep}{7pt}
\footnotesize
\begin{tabular}{@{}lcc@{}}
\toprule
\textbf{Dataset} & \textbf{Mean active / total} & \textbf{Retention (\%)} \\
\midrule
VOC21       & 3.7 / 21   & 97.4 \\
VOC20       & 3.1 / 26   & 99.4 \\
Cityscapes  & 18.0 / 23  & 99.0 \\
Context59   & 12.9 / 59  & 93.4 \\
Context60   & 13.9 / 60  & 93.4 \\
COCO-Object & 7.3 / 81   & 93.9 \\
COCO-Stuff  & 25.7 / 171 & 79.8 \\
ADE20K      & 30.8 / 150 & 93.6 \\
\bottomrule
\end{tabular}
\end{table}

Table~\ref{tab:preview_resolution} varies only the preview resolution while other parameters are fixed. From 336 to 672 pixels, mean mIoU stays within 0.04 while the time per image grows by up to 30\%, so we use 336 pixels, the fastest setting in this range. ActiveSAM outperforms SegEarth-OV3 (63.9 mean mIoU) at every tested resolution, including 224 pixels (65.26). These results show that the preview step gives us the flexibility in controlling the preference on inference time or segmentation accuracy while showing the stability across different preview resolution.

\begin{table}[tbp]
\centering
\caption{\textbf{Preview resolution.} Signatures estimated at a 672-pixel preview are fixed across all rows. mIoU (\%) is reported on eight benchmarks. Time is averaged over VOC21, Context60, COCO-Object, VOC20, and Context59, with the ratio to 336 in parentheses. Bold marks the default setting.}
\label{tab:preview_resolution}
\setlength{\tabcolsep}{4pt}
\footnotesize
\resizebox{0.91\linewidth}{!}{\begin{tabular}{@{}rrrrrrrrrrl@{}}
\toprule
\textbf{Preview (px)} & \textbf{V21} & \textbf{P60} & \textbf{C-O} & \textbf{V20} & \textbf{P59} & \textbf{C-S} & \textbf{City} & \textbf{ADE} & \textbf{Mean} & \textbf{Time (s/img)} \\
\midrule
224 & 83.11 & 56.41 & 70.66 & 96.20 & 60.03 & 43.58 & 72.49 & 39.57 & 65.26 & 0.131 ($0.97\times$) \\
\textbf{336} & 83.44 & 56.57 & 72.64 & 97.08 & 60.45 & 44.88 & 72.94 & 40.24 & \textbf{66.03} & \textbf{0.135} ($1.00\times$) \\
448 & 83.37 & 56.47 & 72.53 & 97.32 & 60.46 & 44.75 & 73.34 & 40.26 & 66.06 & 0.145 ($1.08\times$) \\
560 & 83.39 & 56.51 & 72.62 & 97.37 & 60.49 & 44.81 & 72.98 & 40.23 & 66.05 & 0.158 ($1.17\times$) \\
672 & 83.33 & 56.52 & 72.52 & 97.31 & 60.89 & 44.76 & 73.02 & 40.17 & 66.07 & 0.175 ($1.30\times$) \\
\bottomrule
\end{tabular}}
\end{table}

\section{Qualitative Comparison}
\label{app:qualitative}
\begin{figure}[p]
  \centering
  \includegraphics[width=\linewidth, page=1]{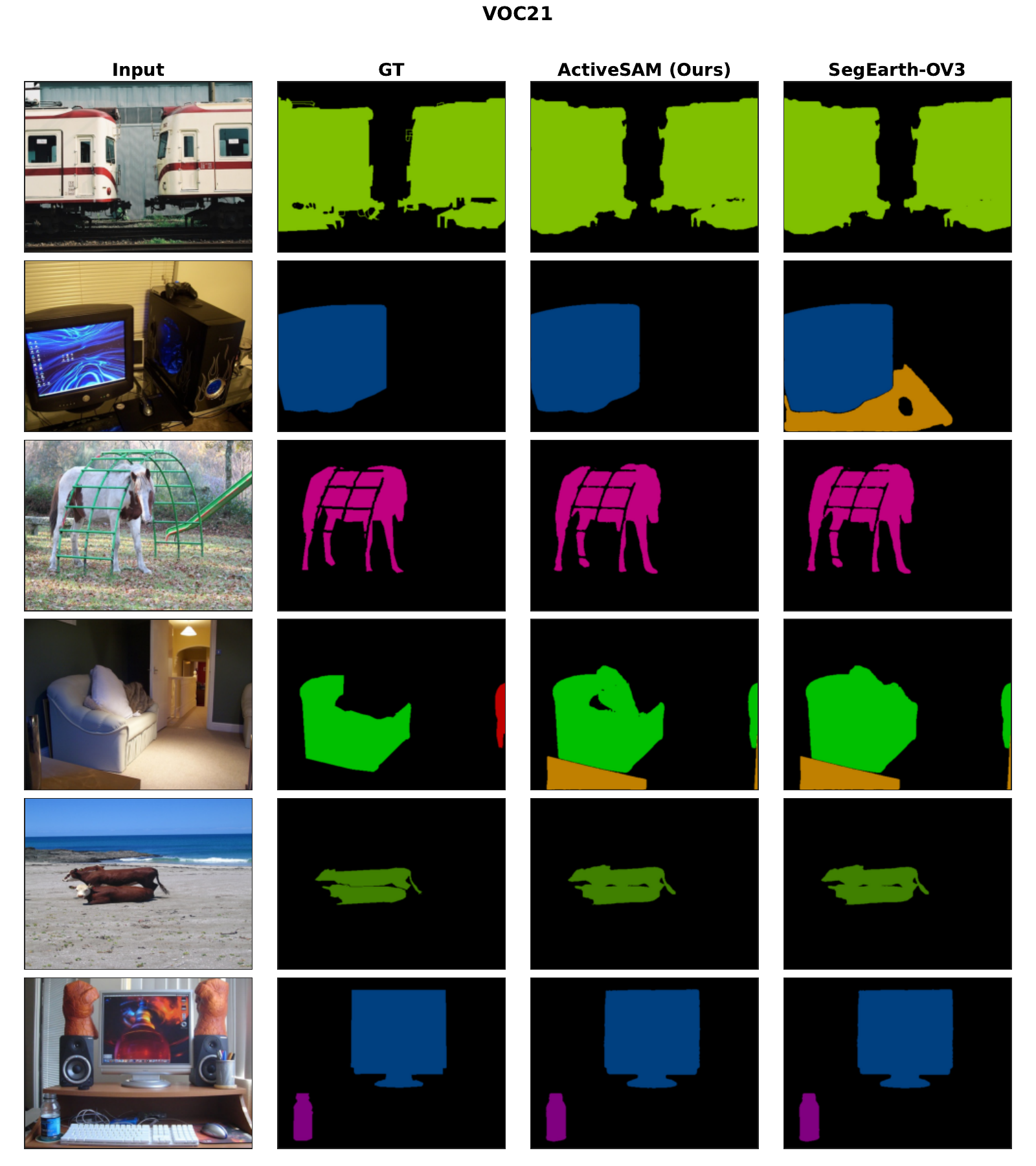}
  \caption{Qualitative results on Pascal VOC21 comparing ActiveSAM (Ours) to SegEarth-OV3 \citep{li2025segearth}.}
  \label{fig:qual_voc21}
\end{figure}

\begin{figure}[p]
  \centering
  \includegraphics[width=\linewidth, page=2]{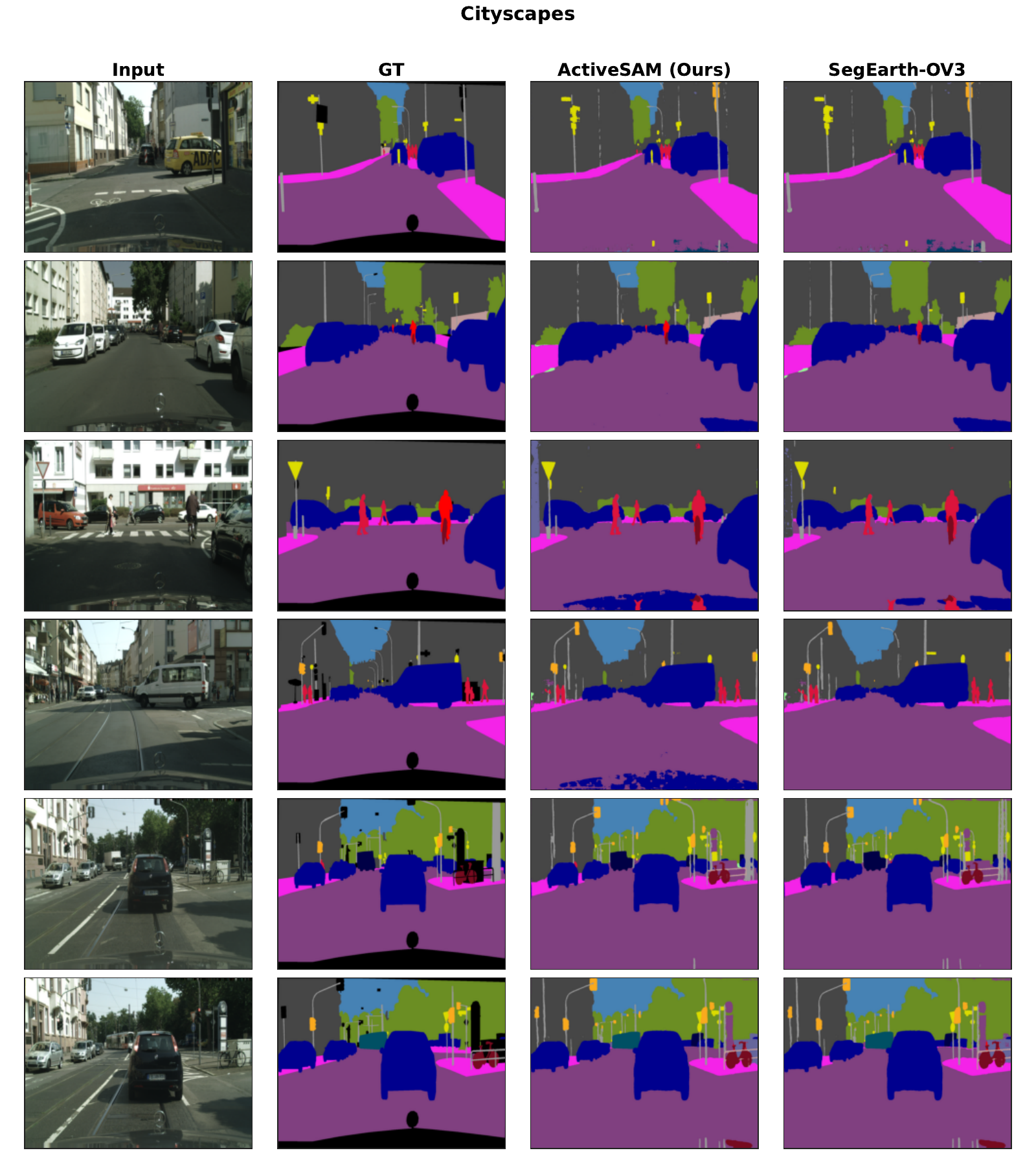}
  \caption{Qualitative results on Cityscapes comparing ActiveSAM (Ours) to SegEarth-OV3 \citep{li2025segearth}.}
  \label{fig:qual_cityscapes}
\end{figure}

\begin{figure}[p]
  \centering
  \includegraphics[width=\linewidth, page=3]{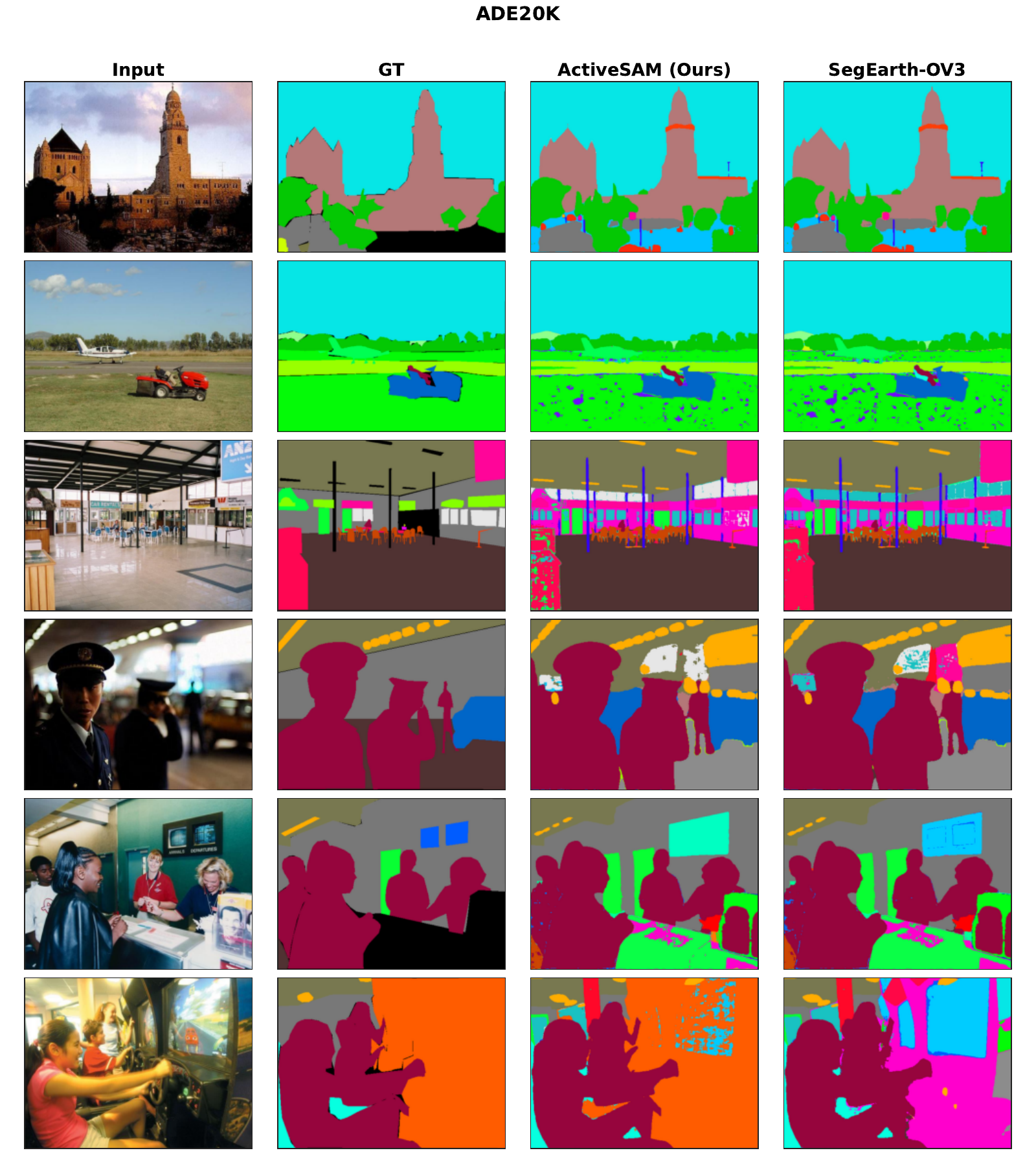}
  \caption{Qualitative results on ADE20K comparing ActiveSAM (Ours) to SegEarth-OV3 \citep{li2025segearth}.}
  \label{fig:qual_ade20k}
\end{figure}

\begin{figure}[p]
  \centering
  \includegraphics[width=\linewidth, page=4]{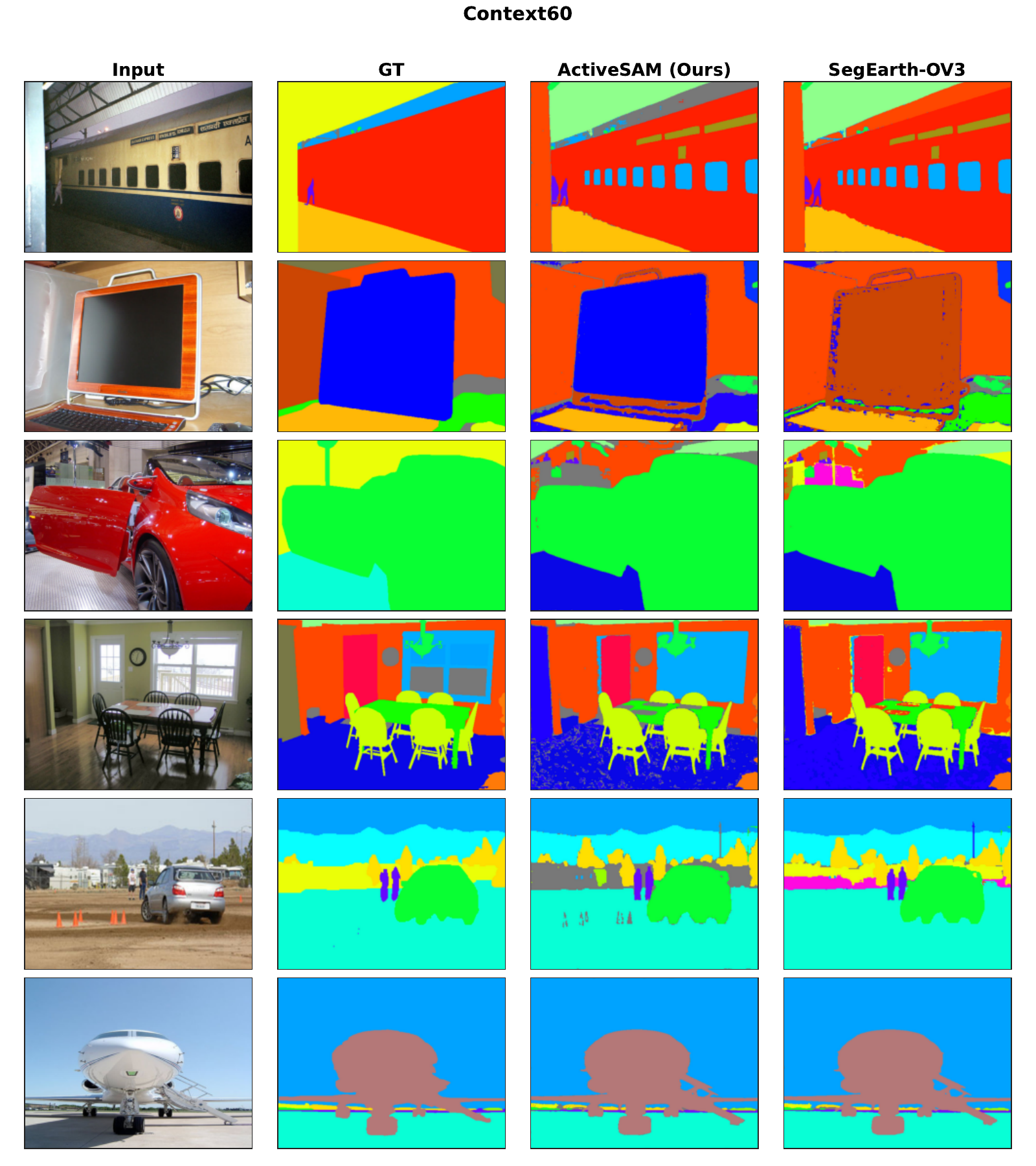}
  \caption{Qualitative results on Context60 comparing ActiveSAM (Ours) to SegEarth-OV3 \citep{li2025segearth}.}
  \label{fig:qual_context60}
\end{figure}

\begin{figure}[p]
  \centering
  \includegraphics[width=\linewidth, page=5]{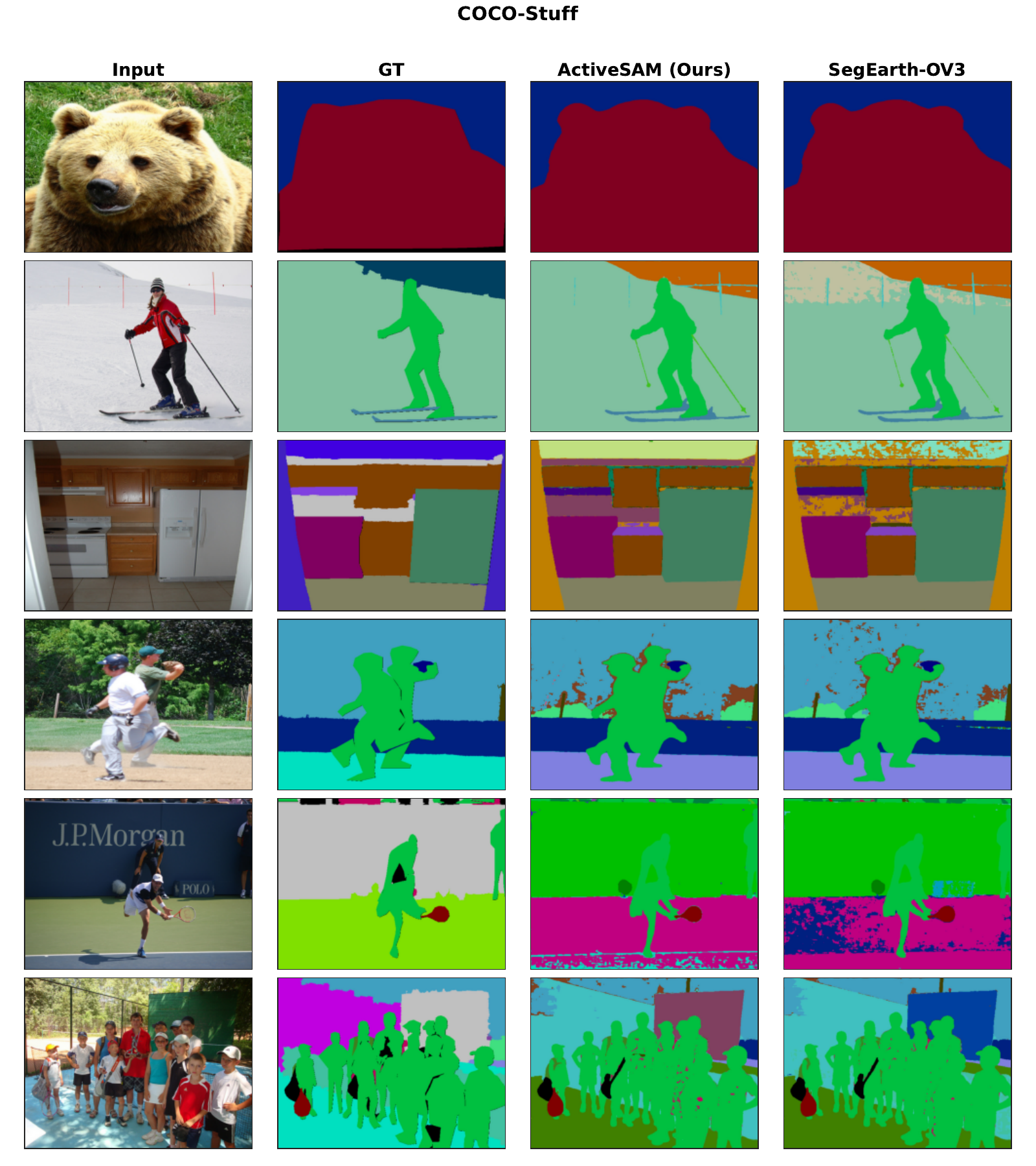}
  \caption{Qualitative results on COCO-Stuff comparing ActiveSAM (Ours) to SegEarth-OV3 \citep{li2025segearth}.}
  \label{fig:qual_cocostuff}
\end{figure}

Figures~\ref{fig:qual_voc21}--\ref{fig:qual_cocostuff} compare the segmentation outputs of ActiveSAM and SegEarth-OV3 \citep{li2025segearth} across five OVSS benchmarks (GT denotes ground truth). For each example, we provide the input image, ground-truth annotations, and the respective predictions from each method. ActiveSAM consistently produces masks with fewer spurious detections, particularly in complex large-vocabulary scenes and background-heavy regions. For instance, ActiveSAM suppresses false positives in the VOC21 monitor scene and the Context60 computer-monitor scene. These visualizations demonstrate that our proposed method yields more robust semantic grounding in diverse environments.

\clearpage 

\clearpage

\clearpage

\clearpage

\clearpage

\end{document}

%% file: math_commands.tex
\usepackage{amsmath,amsfonts,bm}

\def\eqref#1{equation~\ref{#1}}

\def\1{\bm{1}}

\DeclareMathAlphabet{\mathsfit}{\encodingdefault}{\sfdefault}{m}{sl}
\SetMathAlphabet{\mathsfit}{bold}{\encodingdefault}{\sfdefault}{bx}{n}

\DeclareMathOperator*{\argmax}{arg\,max}